\documentclass[10pt,twocolumn,letterpaper]{article}

\usepackage{iccv}
\usepackage{times}
\usepackage{epsfig}
\usepackage{graphicx}
\usepackage{amsmath}
\usepackage{amssymb}
\usepackage{amsmath}
\usepackage{amssymb}
\usepackage{booktabs}
\usepackage{dsfont}
\usepackage{graphicx}
\usepackage{makecell}
\usepackage{multirow}
\usepackage{pifont}
\usepackage[group-separator={,}]{siunitx}
\usepackage{tabularx}
\usepackage{xcolor}
\usepackage{caption}
\usepackage{soul}
\usepackage{changepage,threeparttable}
\usepackage{colortbl}
\usepackage[accsupp]{axessibility}  
\usepackage{mathtools}
\usepackage{xfrac}
\usepackage{subfigure}
\usepackage{catchfile}
\usepackage{longtable}
\usepackage{tabu}
\usepackage{supertabular}
\usepackage{multicol}

\newcolumntype{Y}{>{\centering\arraybackslash}X}

\renewcommand{\ie}{i.e.,}
\renewcommand{\eg}{e.g.,}
\renewcommand{\etal}{\emph{et al.\ }}

\definecolor{color_1}{RGB}{255,0,128}
\definecolor{color_2}{RGB}{0,128,128}
\definecolor{color_3}{RGB}{0,128,0}
\definecolor{color_4}{RGB}{128,0,0}
\definecolor{color_5}{RGB}{128,0,128}
\definecolor{cadetgrey}{RGB}{0.57, 0.64, 0.69}

\newcommand{\salad}[0]{\textsc{SALAD}}
\newcommand{\gaussglot}[0]{\textsc{GaussGlot}}
\newcommand{\Vp}{\mathbf{p}}
\newcommand{\Ve}{\mathbf{e}}
\newcommand{\Vs}{\mathbf{s}}
\newcommand{\Veps}{\boldsymbol{\epsilon}}

\newcommand{\B}{\mathbf}

\usepackage[breaklinks=true,bookmarks=false]{hyperref}
\newif\ifpaper
\papertrue

\iccvfinalcopy 

\def\iccvPaperID{1806} 

\ificcvfinal\fi

\begin{document}

\title{SALAD: Part-Level Latent Diffusion for 3D Shape Generation and Manipulation}


\author{Juil Koo\textsuperscript{$\ast$} $\quad$
Seungwoo Yoo\textsuperscript{$\ast$} $\quad$
Minh Hieu Nguyen\textsuperscript{$\ast$} $\quad$
Minhyuk Sung \\
KAIST\\
{\tt\small \{63days,dreamy1534,hieuristics,mhsung\}@kaist.ac.kr}
}

\twocolumn[{%
\renewcommand\twocolumn[1][]{#1}%
\maketitle
\vspace{-35pt}
\begin{center}
\centering
\captionsetup{type=figure}
\includegraphics[width=\linewidth]{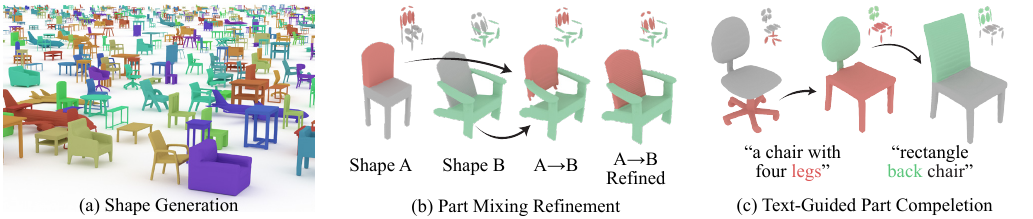}
\vspace{-10pt}
\caption{\textbf{An overview of \salad{}.} (a) Our cascaded diffusion model trained on part-level 3D representations produces high-quality 3D shapes of different classees. Although trained for \emph{unconditional} generation, \salad{} hints its zero-shot capability in various manipulation scenarios, including (b) part mixing and refinement, and (c) text-guided part completion.}

\label{fig:teaser}
\end{center}
}]

\def\thefootnote{*}\footnotetext{Equal contribution.}\def\thefootnote{\arabic{footnote}}

\ificcvfinal\thispagestyle{empty}\fi

\begin{abstract}
\vspace{-5pt}
We present a cascaded diffusion model based on a part-level implicit 3D representation. Our model achieves state-of-the-art generation quality and also enables part-level shape editing and manipulation without any additional training in conditional setup. Diffusion models have demonstrated impressive capabilities in data generation as well as zero-shot completion and editing via a guided reverse process. Recent research on 3D diffusion models has focused on improving their generation capabilities with various data representations, while the absence of structural information has limited their capability in completion and editing tasks. We thus propose our novel diffusion model using a part-level implicit representation. To effectively learn diffusion with high-dimensional embedding vectors of parts, we propose a cascaded framework, learning diffusion first on a low-dimensional subspace encoding extrinsic parameters of parts and then on the other high-dimensional subspace encoding intrinsic attributes. In the experiments, we demonstrate the outperformance of our method compared with the previous ones both in generation and part-level completion and manipulation tasks.
\ificcvfinal
Our project page is \href{https://salad3d.github.io/}{https://salad3d.github.io}.
\fi
\end{abstract}
\vspace{-15pt}
\section{Introduction}
\label{sec:introduction}
\vspace{-5pt}
The staggering rise of the recent image generative model such as DALL-E 2~\cite{Ramesh:2022DALLE2}, StableDiffusion~\cite{Rombach:2022LDM}, and Midjourney~\cite{Midjourney} has drawn great attention to the diffusion models. With the state-of-the-art performance in generating data~\cite{Dhariwal:2021DiffusionBeatGANs, Ho:2019DDPM, Rombach:2022LDM,Ramesh:2022DALLE2,Midjourney}, diffusion models have quickly replaced existing generative models in many applications.
Besides the quality of the generated data, another key advantage of the diffusion models is the zero-shot capability in completion and editing. Recent research~\cite{Chung:2022MCG, Lugmayr:2022Repaint, Meng:2022SDEdit} has shown that diffusion models trained without any conditions can be applied to completion and editing tasks by starting the reverse process from partial data and properly guiding the process.

Such capabilities of the diffusion models have prompted attempts to apply them to 3D generation~\cite{Cai:2020ShapeGF, Luo:2021DPM, Nichol:2022Point-E, Zhou:2021PVD, Zeng:2022LION, Li:2022Diffusion-Sdf, Nam:20223D-LDM, Hui:2022NeuralWavelet}, although likewise the other neural 3D generation and reconstruction work, the key challenge in applying diffusion to 3D is to find an appropriate representation of 3D data. Particularly, to take full advantage of the diffusion models, both producing realistic data and being leveraged to editing and manipulation, a careful design of the 3D data representation is needed.
A naive adaption of the 2D image diffusion models to the 3D voxels is impractical due to the order of magnitude more computation time and memory. Hence, the earlier attempt to apply diffusion or score-based models to 3D (which has also been continued until recently) was to use point clouds as 3D representation~\cite{Cai:2020ShapeGF, Luo:2021DPM, Nichol:2022Point-E}, although the fine details of shapes could not be reproduced since the training computation is still too heavy to increase the resolution --- $2k$ points are used in training.
Later, some hybrid representations have been explored, such as points and voxels~\cite{Zhou:2021PVD}, points and features~\cite{Zeng:2022LION}, voxels and features~\cite{Li:2022Diffusion-Sdf}, although these were still limited in being trained with low-resolution 3D data.
Implicit representation has been proven to be the best to capture fine details in 3D generation and reconstruction~\cite{Park:2019Deepsdf,Chen:2019ImNet,Mescheder:2019OccNet}. Hence, concurrent work~\cite{Li:2022Diffusion-Sdf,Nam:20223D-LDM} introduced latent diffusion methods generating codes that can be decoded into implicit functions of 3D shapes.
However, then the diffusion in a latent space cannot be used for the \emph{guided} reverse process -- \eg~filling a missing part of a shape while preserving the others, and thus the model cannot be exploited for manipulation.
Neural wavelet~\cite{Hui:2022NeuralWavelet} is a notable exception that improves efficiency in training without a latent space but by learning diffusion in spectral wavelet space.
While it succeeded in producing local details, it is still nontrivial to specify a local region to be modified in the spectral space, thus limiting the model to be used in the manipulation tasks.

As a 3D diffusion model feeding two birds with one seed, achieving high-quality \emph{generation} and enabling \emph{manipulation}, we present our novel \textbf{S}hape P\textbf{A}rt-Level \textbf{LA}tent \textbf{D}iffusion Model, dubbed \textbf{\salad\textbf{}}.
Our work is inspired by recent work~\cite{Genova:2020LDIF,Hui:2022NeuralTemplate,Lin:2022Neuform,Hertz:2022Spaghetti} introducing disentangled implicit representations into \emph{parts}. The advantages of the part-level disentangled representation are in the \emph{efficiency} allocating the memory capacity of the latent code effectively to multiple parts, and also in the \emph{locality} allowing each part to be edited independently, thus best fitted to our purpose. We specifically base our work on SPAGHETTI~\cite{Hertz:2022Spaghetti} that learns the part decomposition in a self-supervised way. Each part is described with an independent embedding vector describing the extrinsics and intrinsics of the part as shown in Figure~\ref{fig:teaser}, and thus the parts that need to be edited or replaced can be easily chosen. It is a crucial difference from latent diffusion where the latent codes do not explicitly express any spatial and structural information and voxel diffusion where the region to be modified can only be specified in the 3D space, not in the shape.

Our technical contribution is the diffusion neural network designed to properly handle the characteristics of the part-level implicit representation, which is a \emph{set} of \emph{high-dimensional} embedding vectors. To cope with the set data and achieve permutation invariance while allowing global communications across the parts, we employ Transformer~\cite{Vaswani:2017Attention} and condition each self-attention block with the timestep in the diffusion process. The challenge is also in learning diffusion in the high-dimensional embedding space, which is known to be hard to train~\cite{Yu:2023Video}.
To get around the issue, we introduce a \emph{two-phase cascaded} diffusion model.
We leverage the fact that the part embedding vector is split into a small set of \emph{extrinsic} parameters approximating the shape of a part and a high-dimensional \emph{intrinsic} latent supplementing the detailed geometry information.
Hence, our cascaded pipeline learns two diffusions, one generating extrinsic parameters first and the other producing an intrinsic latent conditioned on the extrinsics, effectively improving the generation quality with the same computation resources.

Our quantitative and qualitative assessments on \salad{} demonstrate its outperformance compared with SotA methods in shape generation as shown in Section~\ref{sec:shape_generation}. We further demonstrate zero-shot manipulation capability of our \salad{}, trained solely for unconditional generation, by conducting extensive experiments on downstream tasks, including part completion (Section~\ref{sec:shape_completion}), part mixing and refinement (Section~\ref{sec:part_mixing}). 
Last but not least, we showcase the versatility of \salad{} in modeling multi-modal distributions such as text-guided generation (Section~\ref{sec:text_conditional_generation}) and completion (Section~\ref{sec:text_driven_manipulation}). To summarize, our contributions are:

\vspace{-5pt}
\begin{itemize}
    \setlength\itemsep{0.08em}
    \item We propose \salad{}, a novel diffusion model capable of generating part-level 3D implicit representations.
 
    \item We propose a \emph{two-phase cascaded} diffusion model, effective for handling high-dimensional latent spaces, that sets a new SotA in shape generation.
    \item We demonstrate the importance of orchestrating diffusion models and part-level implicit representation for the zero-shot capability of \salad{} in shape editing.
    \item We further extend our \salad{} to text-guided generation and editing that can synergize with text-driven part segmentation network.
\end{itemize}

\section{Related Work}
\label{sec:related_work}





\paragraph{3D Generative Models.}

The first 3D generative models are based on GAN, learning a distribution of latents that can be decoded into various 3D representations such as point clouds~\cite{Achlioptas:2018LatentGAN,Valsesia:2019GraphConv,Shu:2019Treegan} and implicit representations~\cite{Kleineberg:2020VoxelGAN,Hao:2020Dualsdf,Chen:2019ImNet,Ibing:2021GridBased,Zheng:2022SdfStylegan}. Later research~\cite{Chan:2022EG3D,Gao:2022Get3d} also proposed to leverage a 2D discriminator in the 3D GAN training while projecting the 3D shape to 2D via differentiable rendering~\cite{Laine:2020NVDiffrast, Mildenhall:2020NeRF}. Autoregressive models for 3D data have also been introduced to produce meshes~\cite{Nash:2020Polygen}, point clouds~\cite{Sun:2020PointGrow}, or (ir)regular feature grids~\cite{Zhang:20223DILG,Yan:2022ShapeFormer}, which have also been extended to handle conditional inputs in the completion~\cite{Yan:2022ShapeFormer} and multi-modal generation~\cite{Mittal:2022Autosdf,Fu:2022Shapecrafter} tasks.
Recent work focused on exploiting the better generation capabilities of diffusion and score-based models.
Cai~\etal~\cite{Cai:2020ShapeGF}, Luo and Hu~\cite{Luo:2021DPM}, and Zhou~\etal~\cite{Zhou:2021PVD} were the first proposing score-based~\cite{Cai:2020ShapeGF} or diffusion-based~\cite{Luo:2021DPM,Zhou:2021PVD} frameworks learning distributions of point clouds. Hui~\etal~\cite{Hui:2022NeuralWavelet} proposed to learn diffusion over wavelet coefficients of truncated signed distance functions. The recent success of latent diffusion models (LDMs)~\cite{Rombach:2022LDM} for 2D images also prompted to develop diffusion models operating on latent vectors of either the entire 3D shapes~\cite{Chou:2022Diffusionsdf, Nam:20223D-LDM} or each point~\cite{Zeng:2022LION}, voxel~\cite{Li:2022Diffusion-Sdf}, and triplane~\cite{Shue:2022TriplaneDiffusion} (note that all of them are \emph{concurrent} work except for LION~\cite{Zeng:2022LION}). Conditional models taking texts~\cite{Nichol:2022Point-E} or multimodal data~\cite{Li:2022Diffusion-Sdf, Cheng:2022SDFusion} are also concurrently introduced with our work.

The advances in 3D generative models have shown significant improvement in the quality of produced shapes, although, in our work, we focus on introducing a more \emph{versatile} 3D generative model that can be used not only for shape generation but also for shape editing and completion \emph{without} any additional training for the conditional setups (yet also achieving the SotA generation results). We aim to fully utilize the manipulation capabilities of the diffusion model with a compact part-level implicit representation of 3D shapes.
\vspace{-10pt}

\paragraph{Part-Level Implicit 3D Representations.}
There is a large body of work exploring part-level 3D decomposition, although most of which focuses on segmenting or abstracting a supervised~\cite{Yi:2016SIGA,Qi:2017Pointnet,Qi:2017Pointnet++,Mo:2019Partnet,Mo:2019StructureNet} and unsupervised~\cite{Tulsiani:2017VolumetricPrimitives,Sun:2019HA,Yang:2021Cubseg,Paschalidou2019:Superquadrics,Chen:2020BspNet,Deng:2020Cvxnet,Paschalidou:2021NeuralParts} ways.
Recent work coupled the part-level structure with the implicit shape representation to enable shape manipulation with the part representation parameters. SIF~\cite{Genova:2019LearningShapeTemplates} and LDIF~\cite{Genova:2020LDIF} first introduced the idea of combining a set of Gaussians in the 3D space to local implicit functions corresponding to each of them. 
NeuralTemplate~\cite{Hui:2022NeuralTemplate} instead used a set of convexes as the part-level extrinsics and connected each of them with a latent vector decoded into a local implicit function. 
SPAGHETTI~\cite{Hertz:2022Spaghetti} employed 3D Gaussians again but trained the network so that the Gaussians can not only approximate the shape but also transform a local region with its mean and covariance parameters.
We base our work on SPAGHETTI and present a framework of learning diffusion on the SPAGHETTI representation.
While SPAGHETTI also provided an auto-decoding-based shape generation pipeline, we demonstrate that our cascaded model diffusing on extrinsics and intrinsics sequentially produces shapes with much better quality while learning the exact data distributions on both spaces.

\vspace{-5pt}
\section{Diffusion Models and Part-Level Shape Representation}
\label{sec:background}

\subsection{Background on Diffusion Models}
\label{sec:background_ddpm}
We first briefly overview the technical background of diffusion models.
Diffusion models~\cite{Ho:2019DDPM} are latent variable models that approximate a data distribution $q (\mathbf{x}^{(0)})$ with a Markov chain, which is also called a \emph{reverse process}:
\vspace{-2pt}
\begin{align}
    p_{\theta} (\mathbf{x}^{(0)}) \coloneqq \int p_{\theta} (\mathbf{x}^{(0:T)}) d \mathbf{x}^{(1:T)},
\end{align}
where $p_{\theta} (\mathbf{x}^{(0:T)}) = p (\mathbf{x}^{(T)}) \Pi_{t=1}^{T} p_{\theta} (\mathbf{x}^{(t-1)} \vert \mathbf{x}^{(t)})$.
Here, $p (\mathbf{x}^{(T)}) = \mathcal{N} (\mathbf{x}^{(T)}; \mathbf{0}, \mathbf{I})$ is the standard normal prior
enabling tractable sampling.

The conditional probabilities $\{ p_{\theta} (\mathbf{x}^{(t-1)} \vert \mathbf{x}^{(t)}) \}_{t=1}^{T}$ are parameterized by a neural network whose weights are denoted by $\theta$.
The weights are optimized through the \emph{forward} diffusion process $q (\mathbf{x}^{(1:t)} \vert \mathbf{x}^{(0)})$ that sequentially adds Gaussian noises to the data $\mathbf{x}^{(0)} \sim q (\mathbf{x}^{(0)})$:
\begin{align}
\begin{gathered}
    q (\mathbf{x}^{(1:t)} \vert \mathbf{x}^{(0)}) \coloneqq \Pi_{s=1}^{t} q (\mathbf{x}^{(s)} \vert \mathbf{x}^{(s-1)}), \\
    \text{where} \,\, q (\mathbf{x}^{(s)} \vert \mathbf{x}^{(s-1)}) \coloneqq \mathcal{N} \left(\mathbf{x}^{(s)}; \sqrt{1 - \beta^{(s)}} \mathbf{x}^{(s-1)}, \beta^{(s)} \mathbf{I} \right),
    \raisetag{35pt}
\end{gathered}
\end{align}
and $\beta^{(s)}$ is an element of a monotonically increasing sequence $\beta^{(1:T)} \in (0, 1]^{T}$.
By choosing Gaussians as forward diffusion kernels, the conditional densities $q (\mathbf{x}^{(t)} \vert \mathbf{x}^{(0)})$ at $t=1,\dots,T$ can be expressed in the closed form:
\begin{align}
    q (\mathbf{x}^{(t)} \vert \mathbf{x}^{(0)}) = \mathcal{N} (\mathbf{x}^{(t)}; \sqrt{\bar{\alpha}^{(t)}} \mathbf{x}^{(0)}, (1 - \bar{\alpha}^{(t)}) \mathbf{I}),
\end{align}
where $\alpha^{(t)} \coloneqq 1 - \beta^{(t)}$ and $\bar{\alpha}^{(t)} \coloneqq \Pi_{s=1}^{t} \alpha^{(s)}$.
Over the forward process dissipating a sample $\mathbf{x}^{(0)} \sim q (\mathbf{x}^{(0)})$ toward $q(\mathbf{x}^{(T)}) = \mathcal{N} (\mathbf{0}, \mathbf{I})$, the weights $\theta$ parameterizing the reverse process $p_{\theta} (\mathbf{x}^{(0)})$ are learned by optimizing the following variational bound on negative log likelihood:
\begin{equation}
\begin{aligned}
    \mathbb{E}_{q (\mathbf{x}^{(0)})} & [-\log p_{\theta} (\mathbf{x}^{(0)})] \leq \\
    &\mathbb{E}_{q (\mathbf{x}^{(0)}, \dots, \mathbf{x}^{(T)})} \left[-\log \frac{p_{\theta} (\mathbf{x}^{(0:T)})}{q (\mathbf{x}^{(1:T)} \vert \mathbf{x}^{(0)})}\right].
\end{aligned}
\end{equation}
Following Ho \etal~\cite{Ho:2019DDPM}, we parameterize our reverse process $p_{\theta} (\mathbf{x}^{(t-1)} \vert \mathbf{x}^{(t)})$ as:
\begin{equation}
\begin{aligned}
    p_{\theta} (\mathbf{x}^{(t-1)} \vert \mathbf{x}^{(t)}) \coloneqq \mathcal{N} (\mathbf{x}^{(t-1)}; \boldsymbol{\mu}_{\theta} (\mathbf{x}^{(t)}, t), \beta^{(t)} \mathbf{I}).
\end{aligned}
\end{equation}
In particular, we use the parameterization $\boldsymbol{\mu}_{\theta} (\mathbf{x}^{(t)}, t) = \sfrac{1}{\sqrt{\alpha^{(t)}}} (\mathbf{x}^{(t)} - \sfrac{\beta^{(t)}}{\sqrt{1 - \bar{\alpha}^{(t)}}} \boldsymbol{\epsilon}_{\theta} (\mathbf{x}^{(t)}, t))$ and optimize its parameters $\theta$ with a training objective that encourages a network $\boldsymbol{\epsilon}_{\theta}$ to predict the noise $\boldsymbol{\epsilon} \sim \mathcal{N}(\mathbf{0}, \mathbf{I})$ present in the given data:
\begin{align}
    \mathcal{L}(\theta) \coloneqq \mathbb{E}_{t, \mathbf{x}^{(0)}, \boldsymbol{\epsilon}} \left[\left\lVert \boldsymbol{\epsilon} - \boldsymbol{\epsilon}_{\theta} \left(\sqrt{\bar{\alpha}^{(t)}} \mathbf{x}^{(0)} + \sqrt{1 - \bar{\alpha}^{(t)}}\boldsymbol{\epsilon}, t\right) \right\rVert^{2}\right].
    \label{eq:ddpm_loss}
\end{align}

\begin{figure}[t!]
\label{fig:spaghetti_overview}
\includegraphics[width=\linewidth]{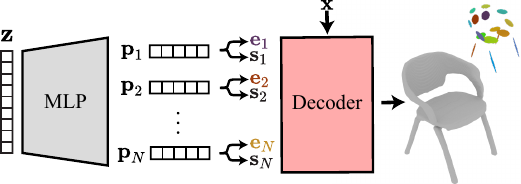}
\caption{\textbf{Part-Level implicit representation by Hertz~\etal~\cite{Hertz:2022Spaghetti}.} A latent vector $\mathbf{z}$ encoding global geometry is first mapped to a set of part latents $\{\mathbf{p}_i\}_{i=1}^N$, each of which is decomposed into extrinsic parameters $\{\mathbf{e}_i\}_{i=1}^N$ and intrinsic latents $\{\mathbf{s}_i\}_{i=1}^N$. The decoder, conditioned on $\{(\mathbf{e}_i, \mathbf{s}_i)\}_{i=1}$, outputs an occupancy value given a query point $\mathbf{x}$.}
\vspace{-10pt}
\end{figure}

\vspace{-15pt}
\subsection{Part-Level Shape Representation}
\label{sec:background_part_representation}
Neural implicit representations~\cite{Chen:2019ImNet, Park:2019Deepsdf, Mescheder:2019OccNet} have been widely exploited in 3D shape generation and reconstruction due to their advantages in capturing fine details without limitation in resolutions even with a small memory footprint. However, their disadvantage of not supporting intuitive editing and manipulation has been a hindrance to increasing their utilization. To remedy the drawback, recent works~\cite{Genova:2019LearningShapeTemplates,Genova:2020LDIF,Hao:2020Dualsdf,Hui:2022NeuralTemplate,Hertz:2022Spaghetti} introduced \emph{dual} representations combining explicit and implicit representations, taking advantage of both of them. Among them, Hertz~\etal~\cite{Hertz:2022Spaghetti}, which our work is based on, was the first introducing a hybrid representation integrating two types of disentanglements simultaneously into an implicit representation: 1) part-level disentanglement, representing each local region separately, and 2) extrinsic-intrinsic disentanglement, describing extrinsic properties (\ie~the approximate shape and transformations) with parameters in the 3D space while encoding intrinsic properties (\ie~geometric details) using a latent code. This novel representation, called SPAGHETTI~\cite{Hertz:2022Spaghetti}, is learned in an auto-decoding setup without any supervision of the part decomposition.

In SPAGHETTI, a 3D shape is first mapped to a global latent $\mathbf{z}$ and then further encoded into a set of part embedding vectors $\{\Vp_i\}_{i=1}^{N}$, where $N$ denotes the number of parts. Each part embedding vector $\Vp_i$ is again mapped into both a set of extrinsic parameters $\Ve_i$ and an intrinsic latent $\Vs_i$ through an MLP. 
The set of extrinsic parameters $\Ve_i = \{\mathbf{c}_i, \mathbf{\Sigma}_i, \mathbf{\pi}_i \}$ of each part represents a Gaussian in the 3D space with mean $\mathbf{c}_i\in\mathbb{R}^3$ and covariance $\mathbf{\Sigma}_i \in \mathbb{R}^{3 \times 3}$, depicting an approximate shape of a part.
$\pi_i \in \mathbb{R}$ is the blending weight for the Gaussian mixture representation of the entire shape: $\sum_{i}\pi_i \mathcal{N}(\B{x} | \B{c}_i, \mathbf{\Sigma}_i)$, describing the volume of the shape as a probability distribution. Since $\{\Ve_i\}_{i=1}^{N}$ can only encode the part-level structural information, the intrinsic latents $\{\Vs_i\}_{i=1}^{N}$ supplement the detailed geometry information so that the pairs of the extrinsic parameters and intrinsic latents can be decoded back to the original shape in an implicit form. Specifically, an implicit decoder $\mathcal{D}$ is trained to predict an occupancy value at point $\mathbf{x}$:
\begin{equation}
\begin{aligned}
\label{eq:decoder}
    o = \mathcal{D}\left(\mathbf{x}\, \Big\vert \, \{\Ve_i\}_{i=1}^{N},\,\{\Vs_i\}_{i=1}^{N}\right),
\end{aligned}
\end{equation}
where occupancy value $o \in [0,1]$ is 1 when the query point is inside the shape, and 0 otherwise. 
The keys to achieving both the part-level and extrinsic-intrinsic disentanglements in the training of decoder $\mathcal{D}$ are the regularizations forcing a single pair $(\Ve_i, \Vs_i)$ of a part to determine the occupancy of each point, and the Gaussian parameters in $\Ve_i$ to transform the corresponding local region. See the original paper~\cite{Hertz:2022Spaghetti} for the details of the decoder training.

The extrinsic vector $\mathbf{e}_i$ is precisely represented as a $16$-dimensional vector $\{\mathbf{c}_i, \lambda_i^1, \lambda_i^2, \lambda_i^3, \mathbf{u}_i^1, \mathbf{u}_i^2, \mathbf{u}_i^3, \pi_i\}$, where $\lambda_i^j \in \mathbb{R}$ and $\mathbf{u}_i^j \in \mathbb{R}^3$ are eigenvalues and eigenvectors of the covariance matrix $\mathbf{\Sigma}_i$, while the intrinsic vector $\mathbf{s}_i$ is a 512-dimensional vector. Note that the much smaller extrinsic vector contains the approximate shape information of the part; we leverage this fact in our effective cascaded diffusion model.

Also, note that SPAGHETTI is trained in an auto-decoding setup while regularizing the global latent code $\mathbf{z} \in \mathbb{R}^{512}$ to follow the unit Gaussian. Thus, the shapes can be simply generated by sampling a latent code $\mathbf{z}$ from the unit Gaussian in the $\mathbf{z}$ space, although we demonstrate that diffusion in the extrinsic and intrinsic embedding spaces can produce much more plausible shapes (Section~\ref{sec:shape_generation}).


 
\section{SALAD -- Part-Level Cascaded Diffusion}
\label{sec:method}

\begin{figure*}
\label{fig:pipeline}
\includegraphics[width=\linewidth]{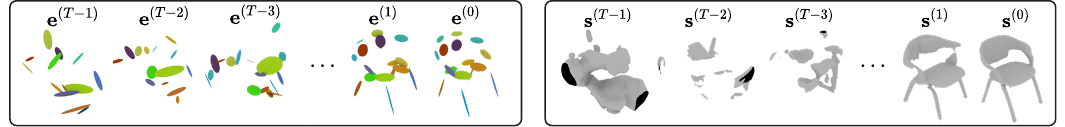}
\caption{\textbf{Pipeline overview.} \salad{} consists of two diffusion models for extrinsic and intrinsic vectors, respectively. During phase 1 (left), it generates extrinsic vectors representing structures of shapes. Phase 2 (right) takes these outputs as conditions and produces intrinsic vectors encoding local geometry information.}
\end{figure*}

\begin{figure*}
\centering
\includegraphics[width=0.8\linewidth]{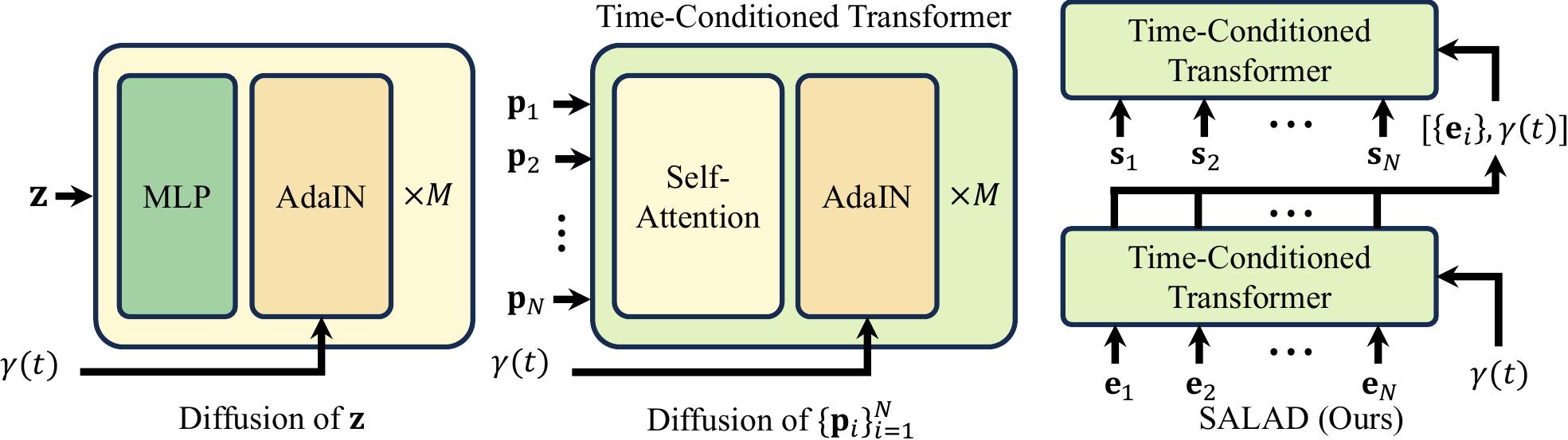}
\caption{\textbf{Architecture diagrams.} The architecture for Diffusion of $\mathbf{z}$ is a sequence of $M$ alternating MLPs and AdaIN~\cite{Perez:2018AdaLN} layers. Time-Conditioned Transformer, a Transformer~\cite{Vaswani:2017Attention} architecture designed to handle diffusion on set data, replaces MLPs with self-attention layers. \salad{} is a cascaded two Time-Conditioned Transformers: one for diffusion of $\{\mathbf{e}_i\}_{i=1}^N$ and the other for $\{\mathbf{s}_i\}_{i=1}^N$. In the second phase of \salad{}, a concatenation of $\{\mathbf{e}_i\}_{i=1}^N$ and $\gamma(t)$ is fed to AdaIN layers as conditioning input.}
\label{fig:pipeline_diagram}
\end{figure*}

Here we introduce our cascaded diffusion framework generating the part-level implicit shape representation.
In the shape representation introduced in Section~\ref{sec:background_part_representation}, note that there are multiple \emph{layers} of representations all of which can be decoded into the original shape, such as the global latent $\mathbf{z}$, the set of part latents $\{\mathbf{p}_i\}$, and the set of extrinsic and intrinsic vectors $\{(\mathbf{e}_i, \mathbf{s}_i)\}$. Below, we first introduce some preliminary approaches to learning diffusion for each representation, and then we propose our final cascaded framework for learning diffusions in two phases.

\vspace{-10pt}
\paragraph{Diffusion of $\mathbf{z}$.}
Learning diffusion in the space of the global shape latent $\mathbf{z}$ is straightforward; the noise prediction network $\boldsymbol{\epsilon}_{\theta}$ (in Equation~\ref{eq:ddpm_loss}) can be simply modeled as an MLP. In the network $\boldsymbol{\epsilon}_{\theta}$, the timestep $t$ is generally first transformed by a positional encoding $\gamma(\cdot)$~\cite{Vaswani:2017Attention} and then fed as the scale and translation factors to the adaptive normalization layers such as AdaIN~\cite{Perez:2018AdaLN}.
In our experiments (Section~\ref{sec:shape_generation}), we show that this simple diffusion already outperforms the quality of generation by sampling $\mathbf{z}$ from the unit Gaussian since it can learn the exact distribution of $\mathbf{z}$, although the improvement is marginal.

\vspace{-10pt}
\paragraph{Diffusion of $\{ \mathbf{p}_i \}_{i=1}^{N}$.} To improve the quality of generation, one can instead consider diffusing the set of part latents $\{ \mathbf{p}_i \}_{i=1}^{N}$. A simple MLP taking the concatenation of the part latents as input, however, results in diffusion in a very high-dimensional space and also does not address the order invariance of the set data. We employ Transformer~\cite{Vaswani:2017Attention} to properly handle the set data while also promoting communications across parts. Each self-attention block is equipped with a post-MLP, where the positional-encoded timestep $\gamma(t)$ is fed to the AdaIN layer. This part-level latent diffusion can better reproduce the details of each part, while it still suffers from the difficulty in diffusing in a high-dimensional latent space.

\vspace{-10pt}
\paragraph{Cascaded Diffusion of $\{ \mathbf{e}_i \}_{i=1}^{N}$ and $\{ \mathbf{s}_i \}_{i=1}^{N}$.}
Inspired by Ho~\etal~\cite{Ho:2022CascadedLDM} introducing \emph{cascaded} diffusion for images, diffusing low-resolution images first and then diffusing high-resolution images conditioned on the low-resolution outputs, we propose a \emph{two-phase} framework for learning diffusion. We observe that the extrinsic and intrinsic attributes $\{ \mathbf{e}_i \}_{i=1}^{N}$ and $\{ \mathbf{s}_i \}_{i=1}^{N}$ play similar roles to low- and high-resolution images; the former describes the approximate of the data, while the latter captures fine details. Also importantly, the extrinsic vector $\mathbf{e}_i$ is much lower-dimensional, thus easier to make the noise prediction converge. Thus, in our first phase, we learn the diffusion of $\{ \mathbf{e}_i \}_{i=1}^{N}$ with the same Transformer-based noise prediction network $\boldsymbol{\epsilon}_{\theta}$ above. Then, in the second phase, we use another Transformer-based network $\boldsymbol{\epsilon}_{\phi}$ to model a conditional distribution $p(\{ \mathbf{s}_i \}_{i=1}^{N} | \{\mathbf{e}_i \}_{i=1}^{N})$ given $\{\mathbf{e}_i \}_{i=1}^{N}$.
Specifically, in the post-MLP of the self-attention block, for each $\mathbf{s}_i$, now the AdaIN layer takes as input a concatenation of the positional-encoded timestep $\gamma(t)$ and a feature vector $\mathcal{E}(\mathbf{e}_i)$ learned from the corresponding extrinsic parameters $\mathbf{e}_i$. The features $\{\mathcal{E}(\mathbf{e}_i) \}_{i=1}^{N}$ are learned from an additional stack of the self-attention modules encoding $\{ \mathbf{e}_i \}_{i=1}^{N}$. Both of the noise prediction networks $\boldsymbol{\epsilon}_{\theta}$ and $\boldsymbol{\epsilon}_{\phi}$ are trained with the same variational bound loss with Equation~\ref{eq:ddpm_loss} as follows:
\begin{equation}
\begin{aligned}
\label{eq:simple_loss}
\vspace{-\baselineskip}
    \mathcal{L}_{\mathbf{e}}(\theta) &:= \mathbb{E}_{t,\{\mathbf{e}\}_{i=1}^N,\Veps} \left[ \left\lVert \Veps - \Veps_{\theta}\left(\{\mathbf{e}^{(t)}\}_{i=1}^N,\gamma(t)\right) \right\rVert^2 \right]
    \raisetag{20pt}
\end{aligned}
\end{equation}
\begin{equation}
\begin{aligned}
\vspace{-\baselineskip}
    \mathcal{L}_{\mathbf{s}}(\phi)&:= \mathbb{E}_{t,\{\mathbf{s}\}_{i=1}^N,\Veps} \left[ \left\lVert \Veps - \Veps_{\phi}\left(\{\mathbf{s}^{(t)}\}_{i=1}^N,\gamma(t), \{\mathbf{e}^{(0)}\}_{i=1}^N\right) \right\rVert^2 \right]
\end{aligned}
\end{equation}
where $\mathbf{e}^{(t)}$ and $\mathbf{s}^{(t)}$ are the extrinsic and intrinsic attributes after $t$-step forward process of adding Gaussian noise, respectively. Refer to the \textbf{supplementary material} for more implementation details.

\section{Experiment}
\label{sec:experiment}
In this section, we demonstrate that~\salad{} outperforms other baselines in shape \emph{generation} (Section~\ref{sec:shape_generation}) and enables intuitive \emph{manipulation}, such as part completion (Section~\ref{sec:shape_completion}) and part mixing and refinement (Section~\ref{sec:part_mixing}), where the combination of part-level representation and diffusion models is essential.
Lastly, we also demonstrate that~\salad{} outperforms other baselines in text-guided shape generation (Section~\ref{sec:text_conditional_generation}) and can leverage part-level representation for text-guided part completion (Section~\ref{sec:text_driven_manipulation}).

\subsection{Shape Generation}
\label{sec:shape_generation}
\paragraph{Evaluation Setup.}
For evaluation and comparison, we follow the settings of Hui~\etal~\cite{Hui:2022NeuralWavelet}. We use \emph{airplane} and \emph{chair} classes from the ShapeNet~\cite{Chang:2015Shapenet} dataset and the train-test split from Chen~\etal~\cite{Chen:2019ImNet}. The model is trained for each class. At inference time, we sample \num{2000} shapes for each class, and measure three evaluation metrics~\cite{Achlioptas:2018LatentGAN,Lopez-paz:20171nna} to assess quality and diversity of the generated shapes:
Coverage (COV), Minimum Matching Distance (MMD), and 1-Nearest Neighbor Accuracy (1-NNA).
We compare \salad{} with existing 3D generative models~\cite{Chen:2019ImNet,Kleineberg:2020VoxelGAN,Luo:2021DPM,Hertz:2022Spaghetti,Hui:2022NeuralWavelet}.


\paragraph{Results.} 
The quantitative and qualitative results, including ablation studies, are summarized in Table~\ref{tbl:quantitative_comparison_of_shape_generation} and Figure~\ref{fig:shape_generation_qualitative_results}.
For more results, refer to the \textbf{supplementary material}.
We reproduced the results of SPAGHETTI~\cite{Hertz:2022Spaghetti} and Neural Wavelet~\cite{Hui:2022NeuralWavelet} using the official code,
and the other quantitative results are directly borrowed from Hui~\etal~\cite{Hui:2022NeuralWavelet}, marked with ``$*$'' in Table~\ref{tbl:quantitative_comparison_of_shape_generation}.
(We also display the results of SPAGHETTI~\cite{Hertz:2022Spaghetti} and Neural Wavelet~\cite{Hui:2022NeuralWavelet} reported by Hui~\etal~\cite{Hui:2022NeuralWavelet}~\cite{Hui:2022NeuralWavelet} in the gray-colored rows. Note that SPAGHETTI results are similar, while there is a gap in the Neural Wavelet results.)
To ease qualitative comparisons in Figure~\ref{fig:shape_generation_qualitative_results},
we retrieve the generated shapes using the same query ground truth shape and compare them.

As shown in Table~\ref{tbl:quantitative_comparison_of_shape_generation},~\salad{} achieves SotA results or is on par with the baselines. In particular, we outperform Neural Wavelet~\cite{Hui:2022NeuralWavelet}, which is a SotA diffusion-based 3D generative model, on 1-NNA by a large margin: \num{65.04} vs. \num{57.82} for \emph{chair} CD, and \num{75.77} vs. \num{73.92} for \emph{airplane} CD (lower is better).

Qualitatively,~\salad{} produces clean high-resolution meshes with fine details as shown in Figure~\ref{fig:shape_generation_qualitative_results}. When comparing ``Diffusion of $\B{z}$'' (in Section~\ref{sec:method}) with SPAGHETTI~\cite{Hertz:2022Spaghetti}, we demonstrate that our simple latent diffusion already produces much better quality shapes than sampling $\B{z}$ from the unit Gaussian distribution as SPAGHETTI does. 
``Diffusion of $\{\B{p}_i\}_{i=1}^N$'' uses Transformer~\cite{Vaswani:2017Attention} instead of simple MLPs and outperforms ``Diffusion of $\B{z}$'', clearly showing how our Transformer-based architecture is the key to learning the distribution of high-dimensional latents represented as a set.

When comparing our final model~\salad{} with ``Diffusion of $\{\B{p}_i\}_{i=1}^N$'',~\salad{} outperforms ``Diffusion of $\{\B{p}_i\}_{i=1}^N$'' by a large margin across all metrics. It shows that our cascaded diffusion training is crucial to improve shape generation quality.



\begin{table*}[t!]
\centering
\newcolumntype{Y}{>{\centering\arraybackslash}X}
\caption{\textbf{Quantitative comparison of shape generation.} The numbers directly from Hui~\etal\cite{Hui:2022NeuralWavelet} are marked with *.
MMD-CD scores and MMD-EMD scores are scaled by $10^3$ and $10^2$, respectively. The best results are highlighted without considering the gray-colored rows. 
The ablation study results are presented in rows 8-9.}
\footnotesize
{
\setlength{\tabcolsep}{0.2em}
\renewcommand{\arraystretch}{1.0}
\definecolor{LightCyan}{rgb}{0.88,1,1}
\definecolor{Gray}{gray}{0.85}
\begin{tabularx}{\linewidth}{>{\centering}m{0.5cm} |>{\centering}m{3.5cm}| Y Y Y Y Y Y | Y Y Y Y Y Y }
  \toprule
  \multirow{3}{*}{Id}       &
  \multirow{3}{*}{Method} & \multicolumn{6}{c|}{Chair} & \multicolumn{6}{c}{Airplane} \\
   &              & \multicolumn{2}{c}{COV $\uparrow$} & \multicolumn{2}{c}{MMD $\downarrow$} & \multicolumn{2}{c|}{1-NNA $\downarrow$}  & \multicolumn{2}{c}{COV $\uparrow$} & \multicolumn{2}{c}{MMD $\downarrow$} & \multicolumn{2}{c}{1-NNA $\downarrow$} \\
   &                      &   CD   &   EMD   &   CD   &   EMD   &   CD   &   EMD   &   CD   &   EMD   &   CD   &   EMD   &   CD   &   EMD \\
  \midrule
  1 & IM-NET$*$~\cite{Chen:2019ImNet}    & \textbf{56.49}  & 54.50   & 11.79  & 14.52   & 61.98  & 63.45   & 61.55  & 62.79   & \textbf{3.320}  & 8.371   & 76.21  & 76.08 \\
  2 & Voxel-GAN$*$~\cite{Kleineberg:2020VoxelGAN}       & 43.95  & 39.45   & 15.18  & 17.32   & 80.27  & 81.16   & 38.44  & 39.18   & 5.937  & 11.69   & 93.14  & 92.77 \\
  3 & DPM$*$~\cite{Luo:2021DPM}      & 51.47  & \textbf{55.97}   & 12.79  & 16.12   & 61.76  & 63.72   & 60.19  & 62.30   & 3.543  & 9.519   & 74.60  & 72.31 \\
  \rowcolor{Gray}
  4 & SPAGHETTI$*$~\cite{Hertz:2022Spaghetti} & 49.19  & 51.92   & 14.90  & 15.90   & 70.72  & 68.95   & 58.34  & 58.38   & 4.062  & 8.887   & 78.24  & 77.01 \\
  \rowcolor{Gray}
  5 & Neural Wavelet$*$~\cite{Hui:2022NeuralWavelet}  & 58.19  & 55.46   & 11.70  & 14.31   & 61.47  & 61.62   & 64.78  & 64.40   & 3.230  & 7.756   & 71.69  & 66.74 \\
  6 & SPAGHETTI & 49.48 &	50.22 &	14.7  &	15.85 &	72.34 &	69.46 &	56.86 &	58.83 &	4.260 &	8.930 &	79.36 &	78.86 \\
  7 & Neural Wavelet   & 49.63 &	50.15 &	12.12 &	14.25 &	65.04 &	62.87 & 60.94 &	59.09 &	3.528 &	\textbf{7.964} & 75.77 &	72.93 \\
  \midrule
  8 & Diff. of $\B{z}$  & 49.71  &	48.75 &	11.71 &	\textbf{14.12} &	62.72 &	61.25 & 54.88 &	59.33 &	3.877 &	8.958 &	82.20 &	80.35\\
  9 & Diff. of $\{\B{p}_i\}_{i=1}^N$ & 50.96	&51.40	&13.57	&15.41	&66.19	&67.04& 58.59	& 61.80	& 4.264	& 9.230	& 78.80	& 76.14\\  
  \midrule 
  10 & \salad{} (Ours)                  & 56.42	& 55.16	& \textbf{11.69}	& 14.29	& \textbf{57.82}	&\textbf{ 58.41}	& \textbf{63.16}	& \textbf{65.39}	& 3.636	& 8.238	& \textbf{73.92}	& \textbf{71.08} \\
  \bottomrule
\end{tabularx}
}
\label{tbl:quantitative_comparison_of_shape_generation}
\end{table*}

\def\arraystretch{0.0}
\begin{figure*}
\centering
{
\scriptsize
\setlength{\tabcolsep}{0em}
\renewcommand\tabularxcolumn[1]{m{#1}}
\newcolumntype{Y}{>{\centering\arraybackslash}X}
\begin{tabularx}{\textwidth}{YYYYY|YYY|YY}
\centering
DPM~\cite{Luo:2021DPM} & PVD~\cite{Zhou:2021PVD} & LION~\cite{Zeng:2022LION} & Voxel-GAN~\cite{Kleineberg:2020VoxelGAN} & \makecell{Neural \\Wavelet\cite{Hui:2022NeuralWavelet}} & \makecell{SPAGHETTI\\\cite{Hertz:2022Spaghetti}} & \makecell{Diff. of\\$\B{z}$} & \makecell{Diff. of\\$\{\B{p}_i\}_{i=1}^N$} & Gaussians & \makecell{\salad{}\\(Ours)} \\
\midrule 
\multicolumn{10}{c}{\includegraphics[width=\textwidth]{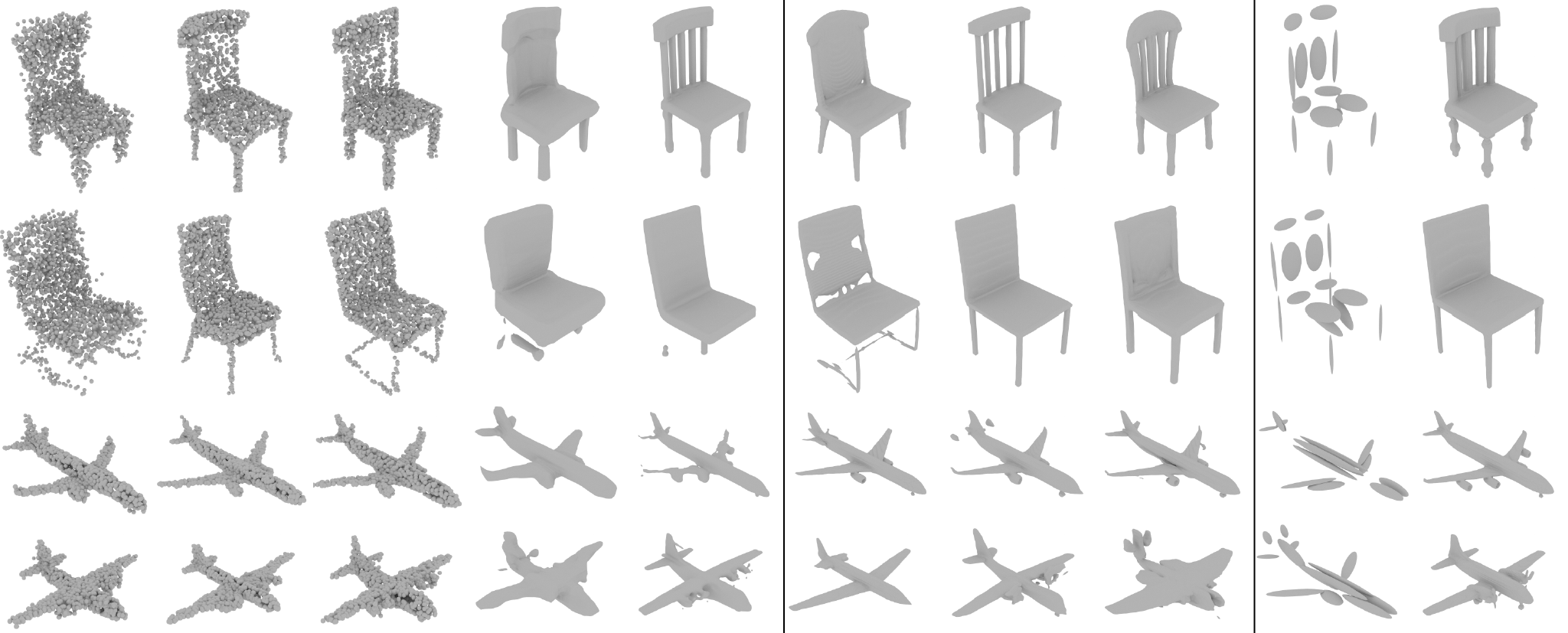}} \\
\end{tabularx}
}
\caption{\textbf{Qualitative comparison of the shape generation.} Given a query ground truth shape, we retrieve the closest generated shape by measuring EMD in each method. \salad{} produces highly detailed 3D shapes compared to the baselines.}
\label{fig:shape_generation_qualitative_results}
\end{figure*}

\subsection{Part Completion}
\label{sec:shape_completion}

\begin{figure*}
\centering
{
\scriptsize
\setlength{\tabcolsep}{0em}
\renewcommand\tabularxcolumn[1]{m{#1}}
\newcolumntype{W}{>{\centering\arraybackslash}m{0.111111\textwidth}}
\newcolumntype{Z}{>{\centering\arraybackslash}m{0.222222\textwidth}}
\begin{tabularx}{\textwidth}{WWW|Z|Z|Z}
GT & Bounding Box & Gaussians & ShapeFormer~\cite{Yan:2022ShapeFormer} &Neural Wavelet~\cite{Hui:2022NeuralWavelet}& {\salad} (Ours) \\ 
\midrule 
\multicolumn{6}{c}{\includegraphics[width=\textwidth]{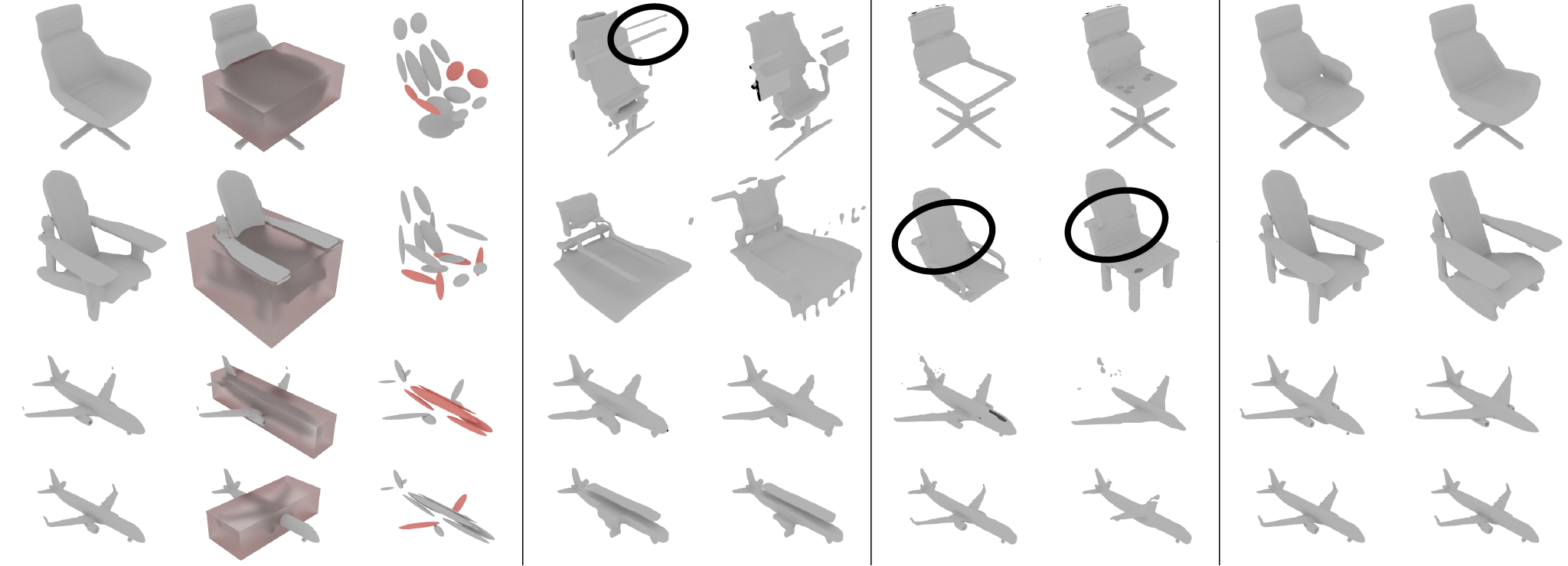}} 
\end{tabularx}
}
\caption{\textbf{Qualitative comparison of the part completion.} We examine \salad{} and other baselines in part completion after ablating semantic parts or regions, highlighted in red in columns 2 and 3. \salad{} produces realistic completions for missing parts. The baselines fail to preserve observed parts or introduce noticeable seams at bounding box boundaries.
}
\label{fig:shape_completion}
\end{figure*}



Here, we describe how~\salad{}, which was trained in an \emph{unconditional} setup, can be employed to part completion. We compare the results against the most recent diffusion model, Neural Wavelet~\cite{Hui:2022NeuralWavelet} and the SotA of shape completion, ShapeFormer~\cite{Yan:2022ShapeFormer}.

\sisetup{group-separator={,}}

\paragraph{Experiment Setup.}
For completion using diffusion models, we run \emph{guided} reverse process proposed by Meng~\etal~\cite{Meng:2022SDEdit}.
Specifically, given the input data $\B{x}\in \mathbb{R}^d$ and a mask of the region to be reconstructed $m \in [0, 1]^d$, each step of the reverse process of the diffusion is performed as follows:
\begin{equation}
 \begin{aligned}
 \B{x}^{(t-1)}_{\text{unmasked}}&\sim\mathcal{N}(\sqrt{\bar\alpha^{(t)}}\B{x}^{(0)},(1-\bar\alpha^{(t)})\B{I}) \\
 \B{x}^{(t-1)}_{\text{masked}}&\sim\mathcal{N}(\boldsymbol{\mu}_\theta(\B{x}^{(t)},t),\beta^{(t)}\B{I}) \\
 \B{x}^{(t-1)}&=m\odot\B{x}^{(t-1)}_{\text{unmasked}}+(1-m)\odot\B{x}^{(t-1)}_{\text{masked}}.
 \end{aligned}
\end{equation}
Unlike previous methods such as ShapeFormer~\cite{Yan:2022ShapeFormer}, this approach guarantees to preserve the unmasked region.
In our experiments, we randomly remove and regenerate a semantic part of \emph{chairs} and \emph{airplanes}. While we can simply select ($\B{e}_i$, $\B{s}_i$) pairs of parts we want to remove in \salad{}, in feature-voxel representation like Neural Wavelet~\cite{Hui:2022NeuralWavelet}, it is not trivial to specify the regions that would include the completed part. This limits their generation output to only occupy the masked voxels, while a larger mask could interfere with or even break unwanted parts leading to seams in the final output. 
For the guided reverse process of Neural Wavelet~\cite{Hui:2022NeuralWavelet} in our experiments, we use the axis-aligned bounding box of a part as a mask and transform the mask to the wavelet domain. Refer to the \textbf{supplementary material} for more details on mask construction.

We first randomly choose 100 shapes from our training set. Then, for all methods, we randomly select a semantic part from each shape and generate five variations.
For quantitative comparisons, we report the reconstruction loss, MMD and FPD (Fréchet PointNet Distance)~\cite{Shu:2019Treegan} indicating the quality and diversity of completions. Note that we measure MMD \emph{from} completions \emph{to} groundtruth shapes to quantify the proximity of the completed shapes to the groundtruth shapes.
We use the official pre-trained models for ShapeFormer~\cite{Yan:2022ShapeFormer} and Neural Wavelet~\cite{Hui:2022NeuralWavelet}. We also report the results from Neural Wavelet trained by ourselves.
\paragraph{Results.}
The quantitative results and qualitative results are summarized in Table~\ref{tbl:part_regeneration} and Figure~\ref{fig:shape_completion}, respectively. 
For more results, refer to the \textbf{supplementary material}.
As shown in Table~\ref{tbl:part_regeneration},~\salad{}, trained solely for \emph{unconditional} shape generation, outperforms the baselines in most of the metrics by large margins, especially in FPD which is the metric of how plausible the shapes are. 


The qualitative results presented in Figure~\ref{fig:shape_completion} further manifests the advantages of employing a part-level 3D representation in \salad{}. In row 1 of Figure~\ref{fig:shape_completion}, ShapeFormer~\cite{Yan:2022ShapeFormer} introduces noticeable artifacts at the back of the chair that lies outside the binary mask (column 2).
In contrast, \salad{} completes the seat seamlessly while preserving the other parts, benefiting from the spatial correspondence between the binary mask and the shape representation. 
Even with such spatial correspondences, the limitation of specifying regions instead of parts persists in Neural Wavelet~\cite{Hui:2022NeuralWavelet}. In particular, the row 2 of Figure~\ref{fig:shape_completion} shows visible seams at the bounding box boundary while \salad{} generates the missing part consistent with the surrounding parts. 

\begin{table*}[t!]
\centering
\newcolumntype{Y}{>{\centering\arraybackslash}X}
\caption{\textbf{Quantitative comparison of part completion.} The metrics based on CD and EMD are scaled by $10^3$ and $10^2$, respectively. The result from the pre-trained Neural Wavelet is marked with *.  
}
{
\footnotesize
\setlength{\tabcolsep}{0.2em}
\renewcommand{\arraystretch}{1.0}
\begin{tabularx}{\linewidth}{>{\centering}m{3.4cm}| Y Y Y Y Y | Y Y Y Y Y}
  \toprule
  \multirow{3}{*}{Method} & \multicolumn{5}{c|}{Chair} & \multicolumn{5}{c}{Airplane} \\
                          & \multicolumn{2}{c}{\emph{reverse}-MMD $\downarrow$} & \multicolumn{2}{c}{Reconstruction $\downarrow$} & \multirow{2}{*}{\makecell{\\FPD $\downarrow$}}  & \multicolumn{2}{c}{\emph{reverse}-MMD $\downarrow$} & \multicolumn{2}{c}{Reconstruction $\downarrow$} & \multirow{2}{*}{\makecell{\\FPD $\downarrow$}} \\  
                          &   CD   &   EMD                       &   CD   &   EMD                                  &                       & CD   &   EMD                       &   CD   &   EMD   & \\
  \midrule
  ShapeFormer~\cite{Yan:2022ShapeFormer} &  32.83 & 22.8 & 55.05 & 25.49 & 83.56 & 5.43 & 10.87 & 10.83 & 11.81 & 79.18 \\
  Neural Wavelet$*$ ~\cite{Hui:2022NeuralWavelet}  & 13.46 & 15.65 & 8.72 & 12.92 & 18.83 & 3.81 & 9.07 & 3.84 & 8.85 & 31.38 \\
  Neural Wavelet & \textbf{11.87} & 15.07 & 8.93 & 12.44 & 18.78 & 3.56 & 8.79 & 3.90 & 9.02 & 36.17 \\
  \midrule
  \salad{} (Ours)                    & 12.1 & \textbf{14.56} & \textbf{5.45} & \textbf{9.22} & \textbf{16.75} & \textbf{3.55} & \textbf{8.68} & \textbf{2.12} & \textbf{6.53} & \textbf{29.44} \\
  \bottomrule
\end{tabularx}
}
\label{tbl:part_regeneration}
\end{table*}

\subsection{Part Mixing and Refinement}
\label{sec:part_mixing}


\begin{figure}
\scriptsize
\setlength{\tabcolsep}{0em}
\renewcommand\tabularxcolumn[1]{m{#1}}
\newcolumntype{Y}{>{\centering\arraybackslash}X}
\begin{tabularx}{\linewidth}{Y Y Y Y}
    Shape A & Shape B & A$\rightarrow$B & \makecell{A$\rightarrow$B\\Refined} \\ 
    \midrule
    \multicolumn{4}{c}{\includegraphics[width=\linewidth]{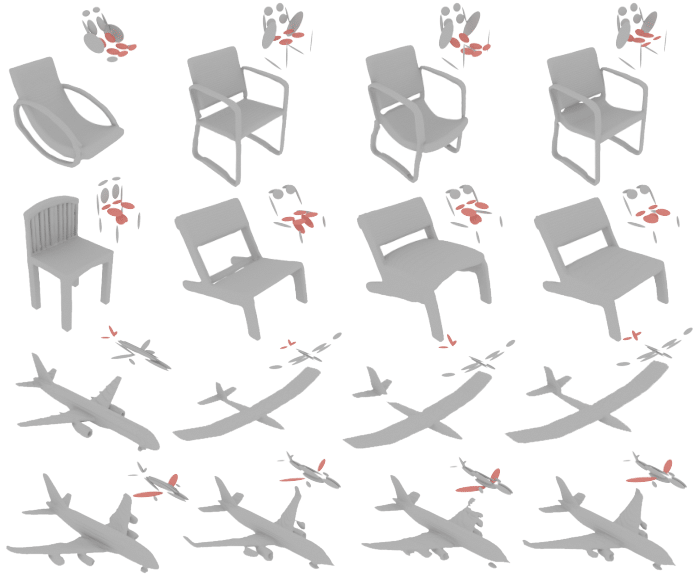}}
    \end{tabularx}
\caption{\textbf{Qualitative results of part mixing and refinement.} \salad{} improves quality of part mixing outputs.}
\label{fig:part_mixing}
\end{figure}



\begin{table*}[h!]
\centering
\newcolumntype{Y}{>{\centering\arraybackslash}X}
\caption{\textbf{Quantitative comparison of part mixing.}
After combining parts from two different shapes, our \salad{} further refines the outputs by adjusting mixed parts. The refinement step brings noticeable improvements in 1-NNA.}
\footnotesize
{
\setlength{\tabcolsep}{0.2em}
\renewcommand{\arraystretch}{1.0}
\definecolor{LightCyan}{rgb}{0.88,1,1}
\definecolor{Gray}{gray}{0.85}
\begin{tabularx}{\linewidth}{>{\centering}m{2.5cm}| Y Y Y Y Y Y | Y Y Y Y Y Y | Y Y Y Y Y Y}
  \toprule
  
  \multirow{3}{*}{Method} & \multicolumn{6}{c|}{Chair} & \multicolumn{6}{c|}{Airplane} & \multicolumn{6}{c}{Table} \\
                 & \multicolumn{2}{c}{COV $\uparrow$} & \multicolumn{2}{c}{MMD $\downarrow$} & \multicolumn{2}{c|}{1-NNA $\downarrow$}  & \multicolumn{2}{c}{COV $\uparrow$} & \multicolumn{2}{c}{MMD $\downarrow$} & \multicolumn{2}{c|}{1-NNA $\downarrow$} & \multicolumn{2}{c}{COV $\uparrow$} & \multicolumn{2}{c}{MMD $\downarrow$} & \multicolumn{2}{c}{1-NNA $\downarrow$} \\
                         &   CD   &   EMD   &   CD   &   EMD   &   CD   &   EMD   &   CD   &   EMD   &   CD   &   EMD   &   CD   &   EMD  &   CD   &   EMD   &   CD   &   EMD   &   CD   &   EMD  \\
  \midrule
   SPAGHETTI~\cite{Hertz:2022Spaghetti} & \textbf{42.24} & \textbf{44.06} & 18.18 & 17.53 & 73.18 & 74.26 & 39.85 & \textbf{42.09} & 5.34 & 10.05 & 80.22 & 78.88 & 31.50 & 32.25 & 19.68 & 18.02 & 86.62 & 87.62 \\
   \salad{} (Ours) & 40.59 & 43.89 & \textbf{17.21} & \textbf{16.96} & \textbf{69.97} & \textbf{68.23} & \textbf{40.15} & 40.75 & \textbf{5.24} & \textbf{9.72} & \textbf{77.61}  & \textbf{76.27} & \textbf{44.25} & \textbf{43.25} & \textbf{17.27} & \textbf{16.98} & \textbf{66.25} & \textbf{69.62} \\
  \bottomrule
\end{tabularx}
}
\label{tbl:quantitative_mixing}
\end{table*}

While Hertz~\etal~\cite{Hertz:2022Spaghetti} demonstrates creating new shapes by combining parts from existing shapes, naively mixing part representations is prone to produce failure cases as illustrated in Figure~\ref{fig:part_mixing} and Figure~\ref{fig:teaser}. Cracks or discontinuities at joint regions are one type of failure case as shown in row 3 of Figure~\ref{fig:part_mixing} and (b) of Figure~\ref{fig:teaser}. Another type of failures is the dissonance between combined parts that results in undesired distortions or the vanishing of parts. 
\salad{} can remedy this issue by refining both the extrinsic and intrinsic vectors through the guided reverse process. Refer to the \textbf{supplementary material} for more qualitative results.

We also show quantitative results of part mixing in Table~\ref{tbl:quantitative_mixing}. 
For evaluation, we use the same metrics and the test set used in Section~\ref{sec:shape_generation}. We randomly select 100 pairs of shapes from the test set and swap a semantic part, for all parts that two shapes in a pair have in common. Swapping a part between two shapes results in two mixed shapes for each pair. The numbers of the shapes resulting from part mixing are 606, 670, and 400 for \emph{chair}, \emph{airplane}, and \emph{table} classes, respectively. The mixed shapes are refined by the guided reverse process with diffusion timestep $t=10$. As indicated in the metrics reported in Table~\ref{tbl:quantitative_mixing}, the quality of mixed shapes are further improved after the refinement step. We particularly observe noticeable gaps in 1-NNA across all shape classes. 
\subsection{Text-Guided Shape Generation}
\label{sec:text_conditional_generation}
We further demonstrate \salad{} can perform \emph{conditional} generation, specifically generating 3D shapes given an input text. To condition a text to the model, we concatenate a language feature and an input of AdaLN, $\gamma(t)$, and optionally $\mathcal{E}(\B{e}_i)$.
We experiment with the text and shape pair dataset from ShapeGlot~\cite{Achlioptas:2019Shapeglot} and compare the generation quality of our text-conditioned model with the one by AutoSDF~\cite{Mittal:2022Autosdf}, which is the SotA text-to-shape generative model. 
The train-test split used in AutoSDF is used.
Also, following AutoSDF, we measure the following three metrics for the evaluation: CLIP-Similarity-Score (CLIP-S)~\cite{Radford:2021CLIP}, Neural-Evaluator-Preference (NEP), and Fréchet Point Cloud Distance (FPD)~\cite{Shu:2019Treegan}.

NEP proposed by Mittal~\etal~\cite{Mittal:2022Autosdf} is a preference rate obtained from a neural evaluator. The neural evaluator is pre-trained on a text-conditioned binary classification task where the model distinguishes the target shape corresponding to the input text. 
Since the neural evaluator used in AutoSDF has not publicly been released, we train our neural evaluator based on PartGlot~\cite{Koo:2022Partglot}, a simpler architecture trained only on point clouds without images. More details of the experiment setup is in the \textbf{supplementary material}. 

\begin{figure}
\scriptsize
\renewcommand\tabularxcolumn[1]{m{#1}}
\newcolumntype{Y}{>{\centering\arraybackslash}X}
\begin{tabularx}{\linewidth}{>{\centering}m{2.7cm}| Y Y| Y Y}
    Text & \multicolumn{2}{c|}{AutoSDF~\cite{Mittal:2022Autosdf}} & \multicolumn{2}{c}{\salad{} (Ours)} \\
    \midrule
    \footnotesize
    \makecell{``\texttt{chair has round}\\\texttt{arms and wheels.}''} &
    \includegraphics[width=0.8\linewidth]{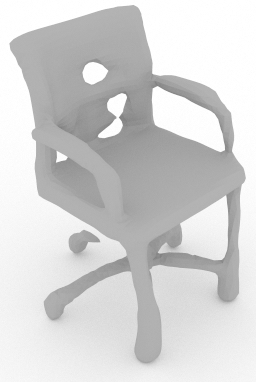} &
    \includegraphics[width=0.8\linewidth]{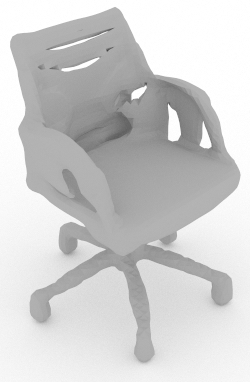} &
    \includegraphics[width=0.8\linewidth]{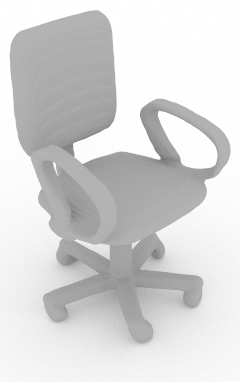} &
    \includegraphics[width=0.8\linewidth]{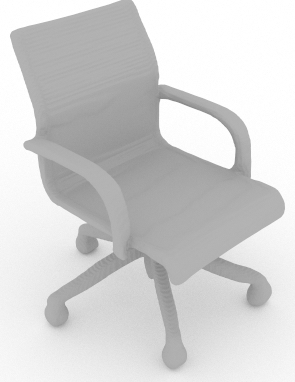} \\ 
    \midrule
    \footnotesize
    \makecell{``\texttt{its the one}\\\texttt{with gaps}\\\texttt{in the back.}''} &
    \includegraphics[width=0.8\linewidth]{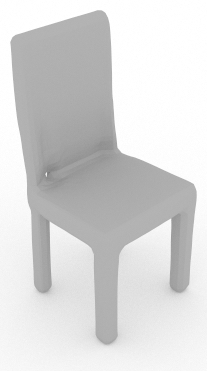} &
    \includegraphics[width=0.8\linewidth]{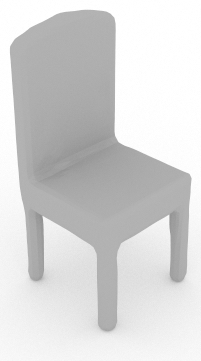} &
    \includegraphics[width=0.8\linewidth]{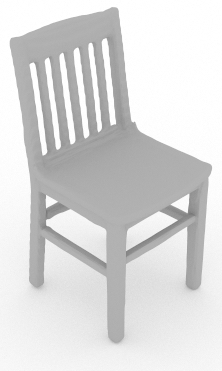} &
    \includegraphics[width=0.8\linewidth]{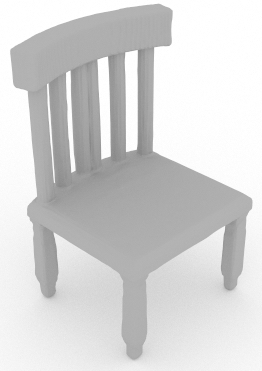} \\
\end{tabularx}
\caption{\textbf{Qualitative comparison of text-guided generation.} \salad{} generates high-quality 3D shapes conforming to the input texts compared to AutoSDF~\cite{Mittal:2022Autosdf}.}
\label{fig:text_generation}
\end{figure}

As shown in Table~\ref{tbl:lang_quantitative_result}, our generated shapes are more preferred by the neural evaluator over the shapes generated by AutoSDF. Also, Figure~\ref{fig:text_generation} and FPD results reflect that ~\salad{} produces more plausible shapes, and our generated shapes conform to given texts more than the shapes of AutoSDF. 


\def\arrvline{\hfil\kern\arraycolsep\vline\kern-\arraycolsep\hfilneg}
\begin{table}[t!]
\centering
\caption{\textbf{Quantitative comparison of text-guided generation.} Overall, \salad{} achieves better performance than AutoSDF. Specifically, it improves FPD by a large margin.}
{
\scriptsize
\setlength{\tabcolsep}{0.2em}
\renewcommand{\arraystretch}{1.0}
\begin{tabularx}{\linewidth}{>{\centering}m{2.3cm}| Y | Y | Y}
\toprule
Methods & CLIP-S $\uparrow$ & NEP $\uparrow$ & FPD $\downarrow$ \\
\midrule 
AutoSDF~\cite{Mittal:2022Autosdf} & \textbf{30.98} & 38.98          & 31.53 \\ 
\salad{} (Ours)                              & 30.92 & \textbf{42.22} &  \textbf{4.043} \\ 

\bottomrule

\end{tabularx}
}
\label{tbl:lang_quantitative_result}
\end{table}

\subsection{Text-Guided Part Completion}
\label{sec:text_driven_manipulation}
We further demonstrate how~\salad{} can be integrated with a text-driven semantic part segmentation network to aid user interactive shape editing. Following PartGlot~\cite{Koo:2022Partglot} architecture, we design \gaussglot{}, a model that uses $\{\B{e}_i\}_{i=1}^N$ as a part representation and predicts semantic part labels of those from texts. More details of \gaussglot{} architecture and training results can be found in the \textbf{supplementary material}. 
Figure~\ref{fig:text_completion} shows examples that the parts of input shapes selected by \gaussglot{} are completed according to given texts by a reverse process of text-conditioned~\salad{} introduced in Section~\ref{sec:text_conditional_generation}. It demonstrates that users can freely manipulate 3D shapes with texts in an end-to-end manner by leveraging~\salad{} with~\gaussglot{}.

\begin{figure}
\footnotesize
\renewcommand\tabularxcolumn[1]{m{#1}}
\newcolumntype{Y}{>{\centering\arraybackslash}X}
\begin{tabularx}{\linewidth}{Y| Y}
    \includegraphics[width=\linewidth]{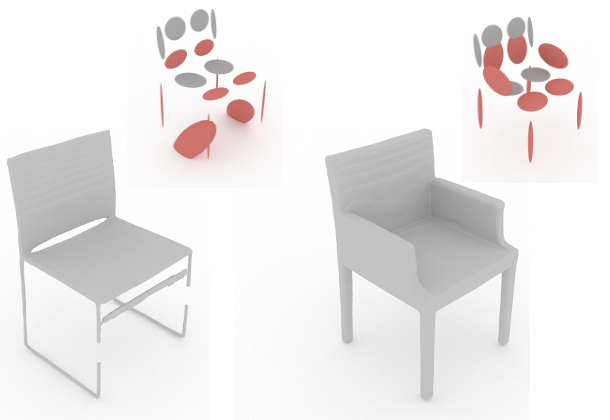} & \includegraphics[width=\linewidth]{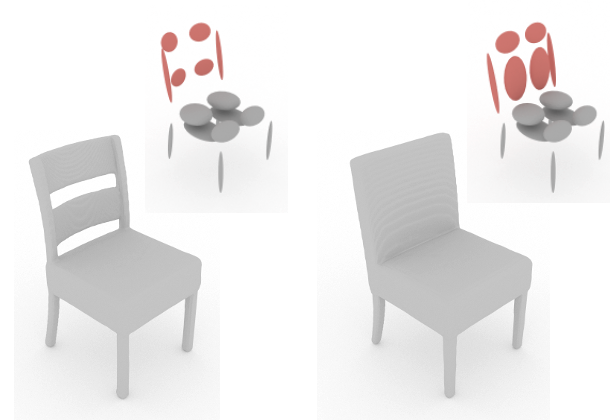} \\ 
    \makecell{``\texttt{four legs and}\\\texttt{two arms.}''} & \makecell{``\texttt{solid back.}''} 
\end{tabularx}
\caption{\textbf{Qualitative results of text-guided part completion.} The part of the left mesh selected by \gaussglot{}, highlighted by red, is completed to fit a given text by a reverse process of text-guided \salad{}.}
\label{fig:text_completion}
\end{figure}



%

\section{Conclusion}
\label{sec:conclusion}

We presented~\salad{}, a cascaded 3D diffusion model for part-level implicit representation. Compared with other 3D diffusion models, our model achieves the best quality in shape generation and also is versatile to be exploited in diverse part-level shape manipulation tasks such as completing, mixing, and text-guided editing. Diffusion on the disentangled representation that allows picking individual parts without specifying a bounding region in the 3D space was the key to fully utilizing the zero-shot manipulation capability of the diffusion models. In future work, we plan to further investigate the diffusion models on part-level representations with different primitives and parametrization for parts.

\vspace{-0.5\baselineskip}
\paragraph{Acknowledgments}
We appreciate Amir Hertz and Ka-Hei Hui for their valuable support in our quantitative comparisons by providing code, data and experimental details.
This work was partly supported by NRF grant (RS-2023-00209723) and IITP grant (2022-0-00594, RS-2023-00227592) funded by the Korean government (MSIT), and grants from ETRI, KT, NCSOFT, and Samsung Electronics.

{\small
\bibliographystyle{ieee_fullname}
\bibliography{egbib}
}

\ifpaper
    \clearpage
    \newpage
    
    \renewcommand{\thesection}{A}
    \renewcommand{\thetable}{A\arabic{table}}
    \renewcommand{\thefigure}{A\arabic{figure}}
    
    \onecolumn
    \section*{Appendix}
\ifpaper
  \newcommand{\refofpaper}[1]{\unskip}
  \newcommand{\refinpaper}[1]{\unskip}
  \newcommand{\suppSegDir}{supp_segmentations}
\else
  \makeatletter
  \newcommand{\manuallabel}[2]{\def\@currentlabel{#2}\label{#1}}
  \makeatother
  \manuallabel{sec:introduction}{1}
  \manuallabel{sec:related_work}{2}
  \manuallabel{sec:background_diffusion}{3.1}
  \manuallabel{sec:background_part_representation}{3.2}
  \manuallabel{sec:method}{4}
  \manuallabel{sec:shape_generation}{5.1}
  \manuallabel{sec:shape_completion}{5.2}
  \manuallabel{sec:results_part_mixing}{5.3}
  \manuallabel{sec:text_conditional_generation}{5.4}
  \manuallabel{sec:text_driven_manipulation}{5.5}
  \manuallabel{fig:shape_generation_qualitative_results}{4}
  \manuallabel{fig:shape_completion}{5}
  \manuallabel{fig:part_mixing}{6}
  \manuallabel{fig:text_generation}{7}
  
  \manuallabel{tbl:quantitative_comparison_of_shape_generation}{1}
  \manuallabel{tbl:part_regeneration}{2}
  
  \newcommand{\refofpaper}[1]{of the main paper}
  \newcommand{\refinpaper}[1]{in the main paper}
\fi

\ifpaper 

\else
\tableofcontents
\clearpage
\newpage
\begin{figure}[h!]
\centering
\includegraphics[width=1\textwidth]{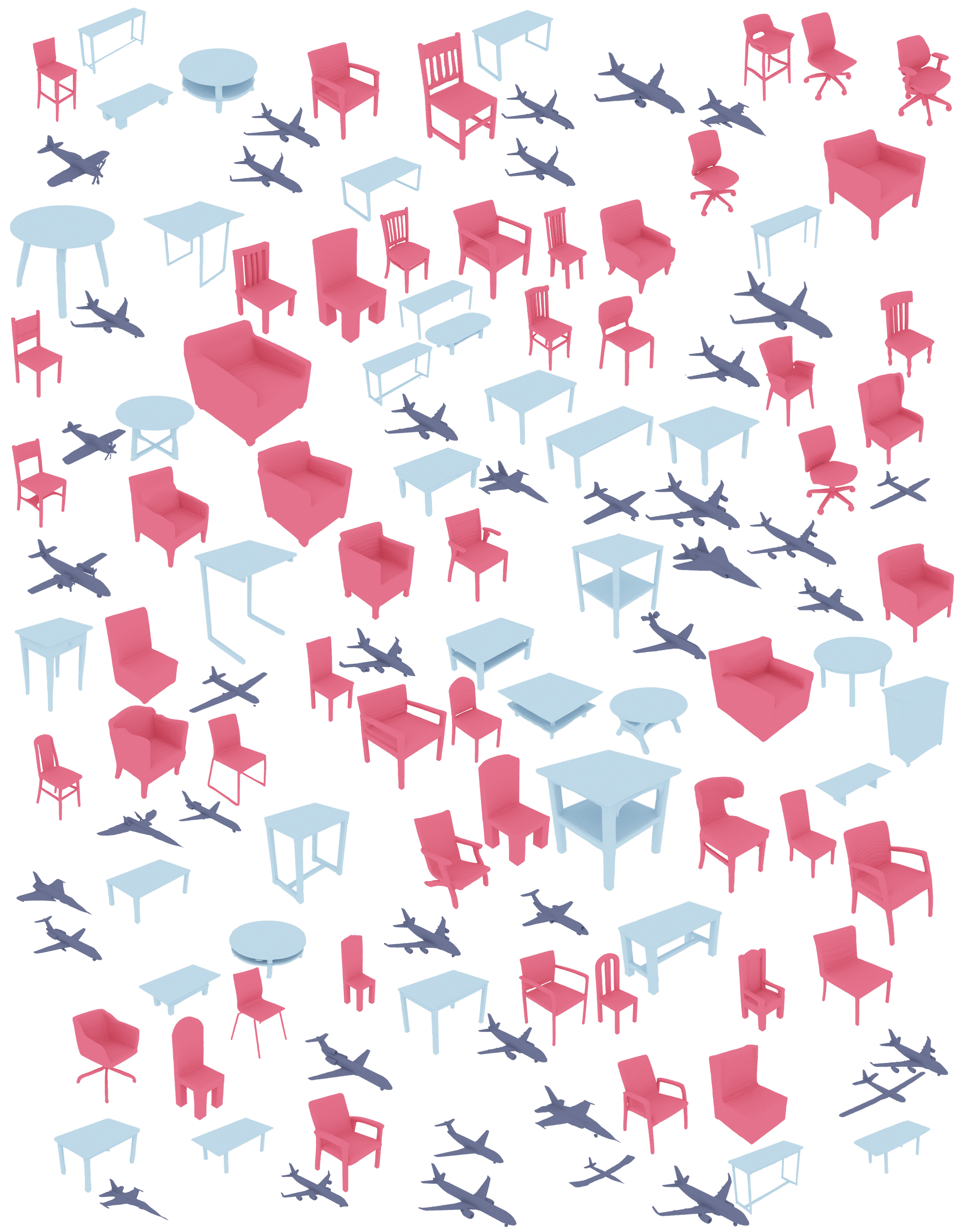}
\caption{\textbf{A visual gallery of \emph{airplanes}, \emph{chairs}, and \emph{tables} generated by~\salad{}.}}
\label{fig:generation_gallery}
\end{figure}
\clearpage
\newpage
\subsection{Overview}
In this supplementary material, we first illustrate additional details in the implementation of \salad{} (Section~\ref{sec:supp_salad_implementation_details}) and details of the experiments discussed in the main paper (Section~\ref{sec:supp_experimental_details}). Then, we report additional experimental results: multi-class generation (Section~\ref{sec:supp_multi_class_generation}), shape generation with more classes (Section~\ref{sec:supp_shape_generation_with_more_classes}) and shape generation with different number of parts (Section~\ref{sec:supp_varying_parts}).
Lastly, we report more \emph{qualitative} results of the experiments reported in the main paper: shape generation (Section~\ref{sec:more_shape_generation_comparison}), part completion (Section~\ref{sec:more_part_completion_comparison}), part mixing and refinement (Section~\ref{sec:more_part_mixing}), text-guided shape generation (Section~\ref{sec:more_text_generation}), and text-guided part completion (Section~\ref{sec:more_text_completion}).

\fi

\subsection{SALAD Implementation Details}
\label{sec:supp_salad_implementation_details}

As discussed in Section~\ref{sec:background_part_representation}~\refofpaper{}, an extrinsic vector $\B{e}_i$ is represented by $\{\mathbf{c}_i, \lambda_i^1, \lambda_i^2, \lambda_i^3, \mathbf{u}_i^1, \mathbf{u}_i^2, \mathbf{u}_i^3, \pi_i\}$, where the eigenvectors $\{\mathbf{u}_i^j\}_{j=1}^{3}$ must be orthogonal to each other. Therefore, the diffusion processes for $\{\B{e}_i\}_{i=1}^N$ need to model distributions in a product space of an orthogonal group $\text{O}(3)$ and Euclidean group, not in the Euclidean space. Recent work~\cite{Leach:2022SO3Diff,Bortoli:2022Riemannian} introduce diffusion models on Lie group or its product space, however, we empirically find that learning diffusion without considering the orthogonality also performs well. It is ensured only at the test time by taking the projection of the generated eigenvectors $\mathbf{U}_i = [\mathbf{u}_i^1, \mathbf{u}_i^2, \mathbf{u}_i^3]$ to $\text{O}(3)$ space. We follow Sch{\"o}nemann~\cite{Schonemann:1966Procrustes} and project $\mathbf{U}_i$ as
\begin{align}
\label{eq:eigvec_projection}
    \tilde{\mathbf{U}}_i = [\tilde{\mathbf{u}}_i^1, \tilde{\mathbf{u}}_i^2, \tilde{\mathbf{u}}_i^3] = \mathbf{A}\mathbf{B}^l,
\end{align}
where $\mathbf{U}_i = \mathbf{A} \boldsymbol{\Sigma} \mathbf{B}^T$ is a singular value decomposition of $\mathbf{U}_i$.
We also clip negative eigenvalues in $\{\lambda_i^j\}_{j=1}^{3}$ to \num{1e-4} since the covariance matrix is positive-definite.

We normalize elements of $\mathbf{e}_i$ to avoid arbitrary high-variance latent space. Specifically, during the training of ``Diffusion of $\{\B{e}_i\}_{i=1}^N$'', we normalize $\pi_i$ and $\{\lambda_i^j\}_{j=1}^3$ using element-wise means and standard deviations pre-computed from all training data. At test time, we re-scale these elements by the means and the standard deviations. We do not apply normalization to the others.

\sisetup{group-separator={,}}
The Transformer-based network of \salad{} introduced in Section~\ref{sec:method}~\refofpaper{} consists of an embedding layer, which maps an input to 512-dimensional embeddings, and 6 Transformer blocks. Each Transformer block is a stack of a self-attention block and an MLP, each of which is followed by an AdaLN layer. We set the dimension of the output of the positional encoding $\gamma(\cdot)$ to 128.

As \salad{} consists of two diffusion models, each trained for \num{5000} epochs, we train the baselines for \num{10000} epochs for a fair comparison. We use a batch size of 64 and an initial learning rate $10^{-4}$ with a polynomial decaying scheduler (power=0.999). The diffusion process is configured with $T=\num{1000}$, $\beta^{(1)}=10^{-4}$, and $\beta^{(T)}=0.05$.


\subsection{Experiment Details}
\label{sec:supp_experimental_details}
In this section, we provide details of the experiments whose results are reported in the main paper.

\subsubsection{Details on Part Completion Experiment Setup --- Section~\ref{sec:shape_completion}}
\label{sec:supp_part_completion_details}
 As mentioned in Section~\ref{sec:shape_completion}~\refofpaper{}, part completion via a \emph{guided} reverse process~\cite{Meng:2022SDEdit} requires binary masks indicating the parts to be ablated. We describe how such masks are constructed for \salad{} and Neural Wavelet~\cite{Hui:2022NeuralWavelet} in this section.

\paragraph{SALAD.}
We define a binary mask $m \in \{0,1\}^N$ for pairs $\{(\mathbf{e}_i, \mathbf{s}_i)\}_{i=1}^N$ to have value \num{0} at completed parts, \num{1} otherwise. To this end, we first \emph{transfer} the part labels of the annotated point clouds from ShapeNet~\cite{Chang:2015Shapenet} dataset to each $(\mathbf{e}_i, \mathbf{s}_i)$. Assume a point cloud $\{ (\mathbf{x}_j, l_j) \}_{j=1}^{K}$ of $K$ points where $\mathbf{x}_j \in \mathbb{R}^3$ and $l_j \in \{1,2,\dots,L\}$, denote 3D coordinate and part label of $j$-th point, respectively. Each $(\mathbf{e}_i, \mathbf{s}_i)$ is assigned a part label $l_i \in \{1,2,\dots,L\}$ based on the proximity of $\mathbf{e}_i$ to the points $\{\mathbf{x}_j\}_{j=1}^K$. Since $\mathbf{e}_i$ parameterizes a Gaussian distribution in 3D space, we employ Mahalanobis distance~\cite{Mahalanobis:1936MahalanobisDistance} as a distance measure. For each Gaussian represented by $\mathbf{e}_i$, we compute the distance to every point $\mathbf{x}_j$ and select the closest \num{100} points. We then count the number of part label occurrences over the points and assign the most frequently occurred label to the pair.

Having assigned the part labels to each of $\{(\mathbf{e}_i, \mathbf{s}_i)\}_{i=1}^N$, we define a mask $m$ selecting a part whose label is $l$ as
\begin{align}
    m_i &= \begin{cases}
        0 & \text{ if } l_i = l \\
        1 & \text{ otherwise} \\
    \end{cases},
\end{align}
where $m_i$ denotes the $i$-th element of $m$.

\paragraph{Neural Wavelet~\cite{Hui:2022NeuralWavelet}.}
Note that there is neither a publicly available official code nor detailed instructions for shape manipulation using Neural Wavelet~\cite{Hui:2022NeuralWavelet}. Although a concurrent work of ours, Hu~\etal~\cite{Hu:2023NeuralWavelet}, demonstrates shape manipulation using Neural Wavelet, it does not provide a detailed implementation.

Following Hui~\etal~\cite{Hui:2022NeuralWavelet}, we derive the wavelet coefficients of the shapes in our training set. We compute signed distance functions (SDFs) of the shapes and truncate their values into $[-0.1, 0.1]$. We denote $S$ the resulting truncated signed distance function (TSDF) of a shape.
We leverage Biorthogonal wavelet-6-8 filter~\cite{Cohen1993:Biorthogonal} to decompose $S$ into a coarse wavelet coefficient volume at a scale 3 ($C^3$) and a detail wavelet coefficient volume at a scale 2 ($D^2$). Refer to Hui~\etal~\cite{Hui:2022NeuralWavelet} for details on preprocessing.

We then aim to derive binary masks for $C^3$, necessary for leveraging pre-trained Neural Wavelet~\cite{Hui:2022NeuralWavelet} for part completion.
Note that selecting a part to complete is a \emph{nontrivial} task for a voxel-based representation adapted by Neural Wavelet, as opposed to \salad{} where we can define binary masks for $\{(\mathbf{e}_i, \mathbf{s}_i)\}_{i=1}^N$ to select parts directly. As one solution, we compute bounding boxes enclosing semantic parts of 3D shapes, and use them to designate the \emph{regions} to complete.
Such bounding boxes are used to compute binary masks for $C^3$ via a heuristic based on the property of wavelet transforms extracting local spectral information.
Through experiments, we empirically find a set of wavelet coefficients that vary when the TSDF values in a 3D volume are set to \num{0.1} (\ie outside of a shape). For instance, we set the TSDF values in the bounding box enclosing the back of a chair to \num{0.1} to discover a set of wavelet coefficients corresponding to the part. We assign \num{0} to the coefficients whose amount of change is above a threshold $\delta$ and \num{1} to the others.

Rigorously, let $M \in \{0,1\}^{256^{3}}$ denote a binary voxel grid of the same resolution as $S$ with 0 indicating the semantic part of interest and 1 otherwise. Such $M$ is derived from a bounding box enclosing a semantic part of a 3D shape, and is used to derive a \emph{masked} TSDF $S^*$ defined as
\begin{equation}
S_v^* = \begin{cases}
   0.1 & \text{ if } M_v = 0  \\ 
   S_v & \text{ otherwise}
\end{cases},
\label{eq:tsdf_mask}
\end{equation}
for all $v \in \{(0,0,0), (0,0,1), ..., (255,255,255)\}$. After marking all values inside a bounding box as \emph{outside}, we obtain the wavelet coefficients $C^{3*}$ via forward wavelet transform. A mask $m$ for $C^{3}$ is then defined as
\begin{equation}
m_{v^{\prime}} =
\begin{cases}
     0 & \text{ if } |C^{3*}_{v^{\prime}} - C^3_{v^{\prime}}| > \delta\\ 
     1 & \text{ otherwise}
\end{cases}
\label{eq:wavelet_mask}.
\end{equation}
for all $v^\prime \in \{ (0,0,0), (0,0,1), ..., (47,47,47)\}$. Here, we use $\delta=0.001$.

\paragraph{ShapeFormer~\cite{Yan:2022ShapeFormer}.}
As discussed in Section~\ref{sec:shape_completion}~\refofpaper{}, after constructing the axis-aligned bounding box of a part, we make a partial point cloud by masking out the points inside the bounding box, and pass it to ShapeFormer~\cite{Yan:2022ShapeFormer} as an input.

\subsubsection{Details on Text-Guided Shape Generation --- Section~\ref{sec:text_conditional_generation}}
\label{sec:supp_text_generation_details}
\paragraph{Implementation Details of Text-Conditioned SALAD.}
We impose text conditions on both the first and the second phase models by feeding text features from our text encoder. We use LSTM~\cite{Hochreiter:1997lstm} for the text encoder and train it jointly with the first and the second phase models. We also apply the classifier-free guidance~\cite{Ho:2021classifierfree}. More precisely, we jointly train a conditional diffusion model $\Veps_\theta(\B{x}^{(t)},t,\B{c})$ and an unconditional diffusion model $\Veps_\theta(\B{x}^{(t)},t,\boldsymbol{\emptyset})$, where $\B{c}$ denotes a condition feature vector and $\boldsymbol{\emptyset}$ is a null condition vector. We randomly set $\B{c}$ to $\boldsymbol{\emptyset}$ with a 20\% dropout probability during training. To make $\boldsymbol{\emptyset}$, we feed an empty sequence as an input text and zero vectors for $\{\mathcal{E}(\B{e}_i)\}_{i=1}^N$. $\B{c}$ is solely a text feature for the first phase model. For the second phase model conditioned on the features from extrinsic vectors $\{\mathbf{e}_i\}_{i=1}^N$, we use the concatenation of the features and a text feature as a condition. 


At sampling time, the noise prediction is adjusted by an extrapolation between the noise prediction of the conditional diffusion model and the unconditional diffusion model as follows:

\begin{equation}
    \tilde{\Veps}_t = (1+w)\Veps_\theta(\B{x}^{(t)},t,\B{c})-w \Veps_\theta(\B{x}^{(t)}, t, \boldsymbol{\emptyset}),
\end{equation}
where $\tilde{\Veps}_t$ is the noise prediction with the classifier-free guidance applied, and $w$ is a hyperparameter controlling guidance strength. We use $w=2$ for sampling.

\paragraph{Experiment Setup.}
To measure Neural-Evaluator-Preference (NEP) discussed in Section~\ref{sec:text_conditional_generation}~\refofpaper{}, we leverage a modified PartGlot~\cite{Koo:2022Partglot} for a neural evaluator. The modified architecture takes point clouds as inputs instead of super-segments. Refer to the PartGlot~\cite{Koo:2022Partglot} paper for more details. 
We adapt the training and test set of PartGlot~\cite{Koo:2022Partglot} to create binary classification examples.
The modified PartGlot achieves $73.98$\% test accuracy on the binary classification. Following Mittal~\etal~\cite{Mittal:2022Autosdf}, we consider an example to be confused if the absolute difference between the neural evaluator's confidence is $\leq 0.2$.

{
\begin{figure}[h!]
\centering
\includegraphics[width=0.6\linewidth]{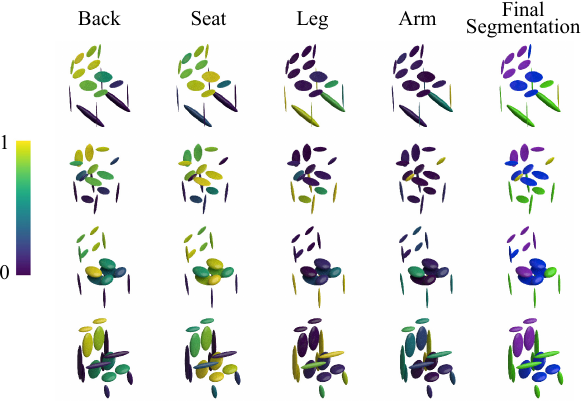}
\caption{\textbf{\gaussglot{} qualitative results.} The attention maps for each semantic part achieved by \gaussglot{} are shown in the left columns of the figure. The colors of the attention maps change from dark blue to yellow as the attention weights increase from 0 to 1. The final part segmentation results are depicted in the rightmost column of the figure, where purple, blue, green, and yellow indicate \emph{back}, \emph{seat}, \emph{leg}, and \emph{arm}, respectively.}

\label{fig:gaussglot_qualitative}
\end{figure}
}

\subsubsection{Details on GaussGlot --- Section~\ref{sec:text_driven_manipulation}}
\label{sec:supp_gaussglot_details}
Inspired by Koo~\etal~\cite{Koo:2022Partglot}, we design a text-driven self-supervised semantic part segmentation network, \gaussglot{}, where a set of Gaussian primitives is employed as super-segments. As discussed in Section~\ref{sec:text_conditional_generation}~\refofpaper{}, PartGlot is a neural evaluator that classifies shapes from a query text. While solving this text-conditioned shape classification, PartGlot learns semantic part segmentation in an unsupervised manner by learning the attention maps between the input text and the super-segments. Refer to the PartGlot~\cite{Koo:2022Partglot} paper for more details.
Specifically, we train \gaussglot{} with $\{\B{e}_i\}_{i=1}^N$ excluding $\pi_i$ elements which is inessential to define 3D Gaussian primitives. Based on the architecture of PartGlot, 15-dimensional Gaussian parameters are mapped to 256-dimensional features through MLPs. We embed text tokens into 128 dimensions and use LSTM as a text encoder with 256-dimensional hidden states. Our trained \gaussglot{} achieves \num{76.03}\% test accuracy and $56.85$\% mIoU. Qualitative part segmentation examples and the attention maps of each semantic part from \gaussglot{} can be found in Figure~\ref{fig:gaussglot_qualitative}.

\subsection{Multi-Class Generation}
\label{sec:supp_multi_class_generation}
\begin{figure}
\centering
\includegraphics[width=0.98\textwidth]{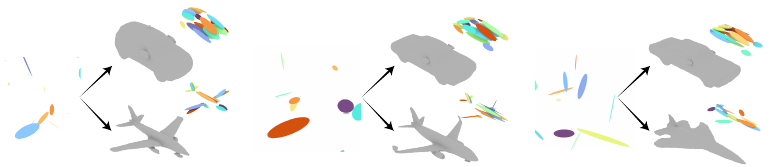}
\caption{\textbf{Class-label-guided generation of \salad{} trained with \emph{airplanes} and \emph{cars}.}}
\label{fig:multi_class_generation}
\end{figure}

We further demonstrate that \salad{} is capable of multi-class generation. We construct the multi-class latent space by pre-training SPAGHETTI~\cite{Hertz:2022Spaghetti} with a training data set consisting of 200 \emph{airplanes} and 200 \emph{cars}. Next, we train class-label-conditioned \salad{} with the latents extracted from the pre-trained SPAGHETTI. Figure~\ref{fig:multi_class_generation} shows the same initial latents are decoded into different class shapes, airplanes and cars, through the class-label-guided reverse process.

\begin{figure}[!htb]
    \begin{minipage}{0.49\textwidth}
     \centering
     \includegraphics[width=.98\linewidth]{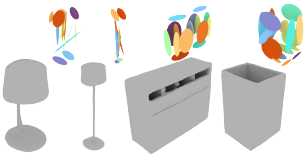}
     \caption{\textbf{Generation of \emph{lamps} and \emph{cabinets}.}}\label{fig:more_class}
   \end{minipage}\hfill
   \begin{minipage}{0.49\textwidth}
     \centering
     \includegraphics[width=.98\linewidth]{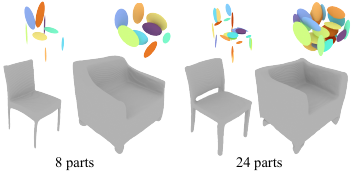}
     \caption{\textbf{Generation with varying number of parts.}}\label{fig:varying_number_parts}
   \end{minipage}
\end{figure}

\subsection{Shape Generation with More Classes}
\label{sec:supp_shape_generation_with_more_classes}


In the main paper, we used \emph{chairs} and \emph{airplanes} for the quantitative comparison as done in the previous work~\cite{Hui:2022NeuralWavelet}. Figure~\ref{fig:multi_class_generation} and Figure~\ref{fig:more_class} show qualitative results of \salad{} trained with more other classes, \emph{cars}, \emph{lamps} and \emph{cabinets}. 

\subsection{Shape Generation with Different Number of Parts}
\label{sec:supp_varying_parts}

Although we used 16 parts in the main paper, Figure~\ref{fig:more_class} shows qualitative results of with varying number of parts, 8 and 24 parts, respectively. It demonstrates that \salad{} is agnostic to the number of parts.
Furthermore, the experiments of Figure~\ref{fig:multi_class_generation}, Figure~\ref{fig:more_class} and Figure~\ref{fig:varying_number_parts} use 400 training shapes, a significantly smaller number than the train set of the main paper. It demonstrates that \salad{} can generate high-quality shapes with a small number of training data.


\ifpaper
\begin{figure}[h!]
\centering
\includegraphics[width=0.8\textwidth]{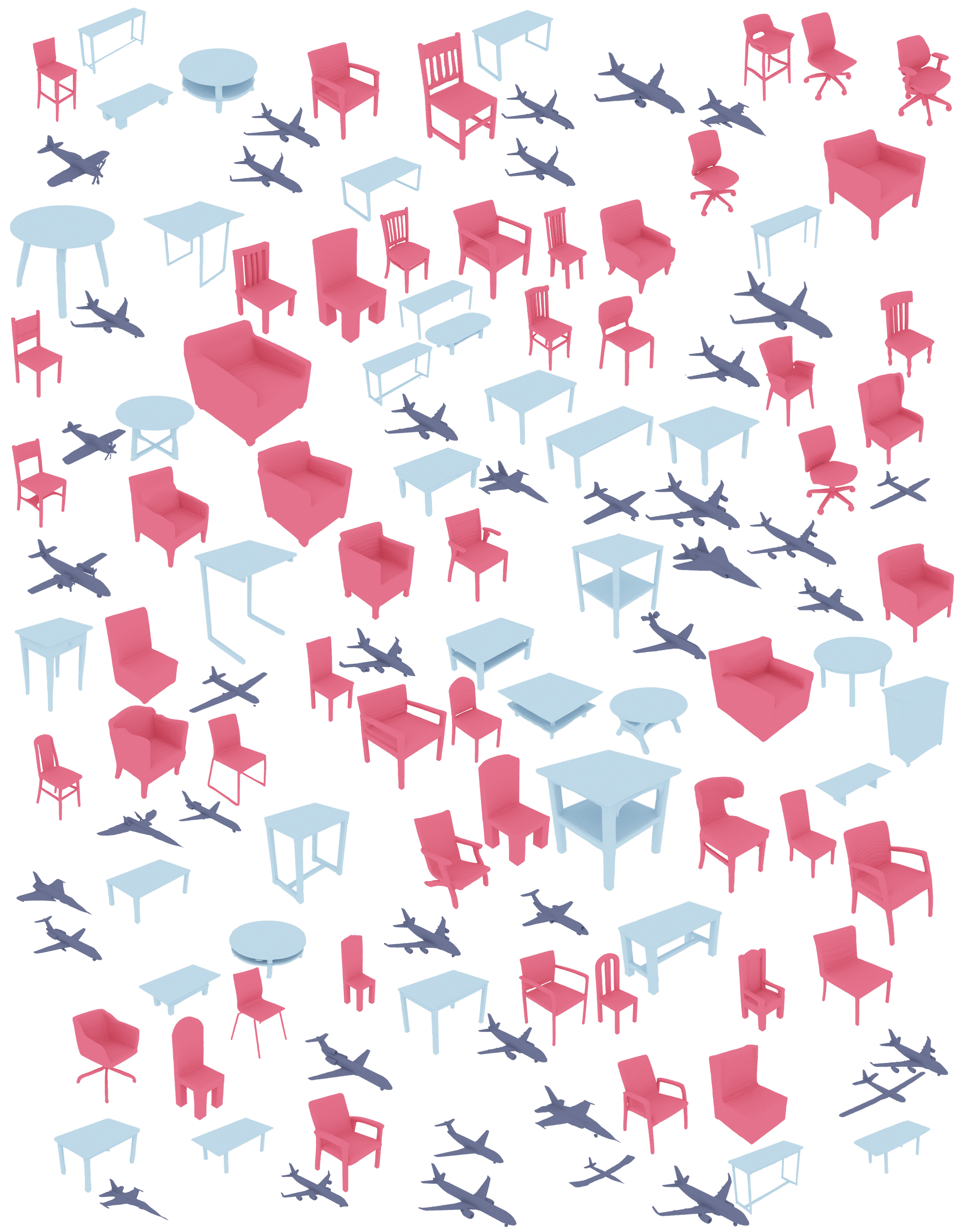}
\caption{\textbf{A visual gallery of \emph{airplanes}, \emph{chairs}, and \emph{tables} generated by~\salad{}.}}
\label{fig:generation_gallery}
\end{figure}
\clearpage
\newpage
\else
\fi 

\clearpage
\newpage
\onecolumn

\subsection{More Qualitative Comparisons on Shape Generation}
\label{sec:more_shape_generation_comparison}

In the following, we provide more qualitative comparisons on shape generation with \emph{chair} and \emph{airplane} classes, as shown in Figure~\ref{fig:shape_generation_qualitative_results}~\refofpaper{}. 
\newcommand{\AllComparisonImages}{\Image{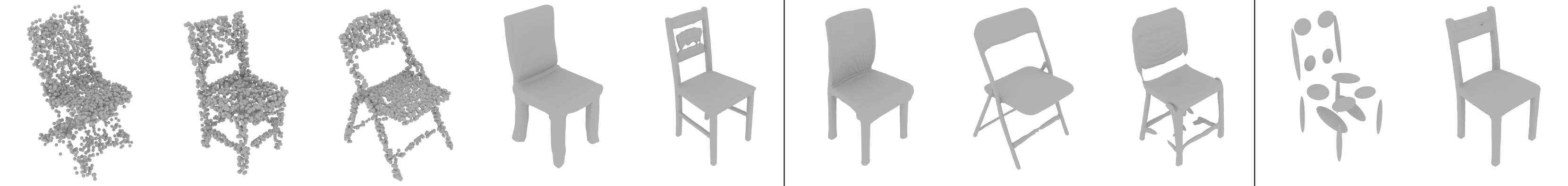}
\Image{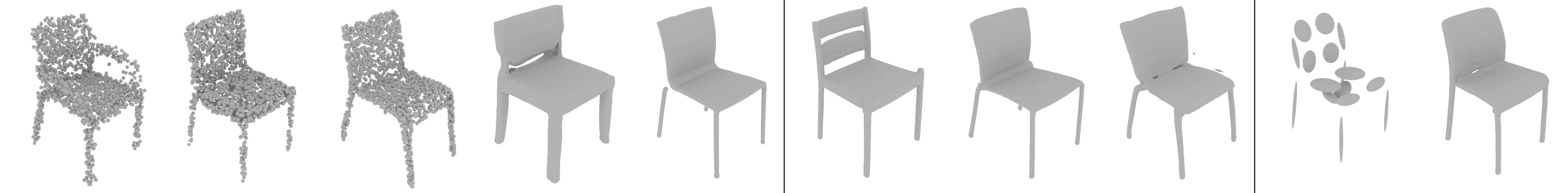}
\Image{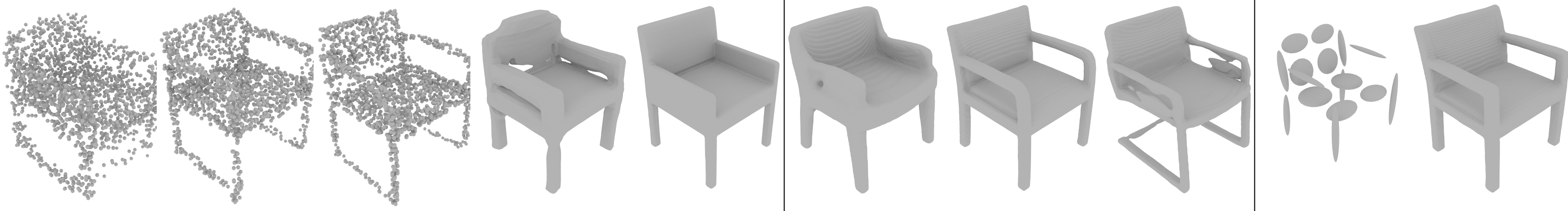}
\Image{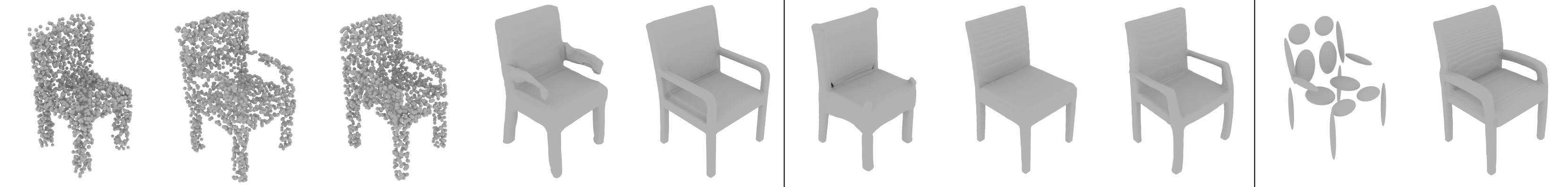}
\Image{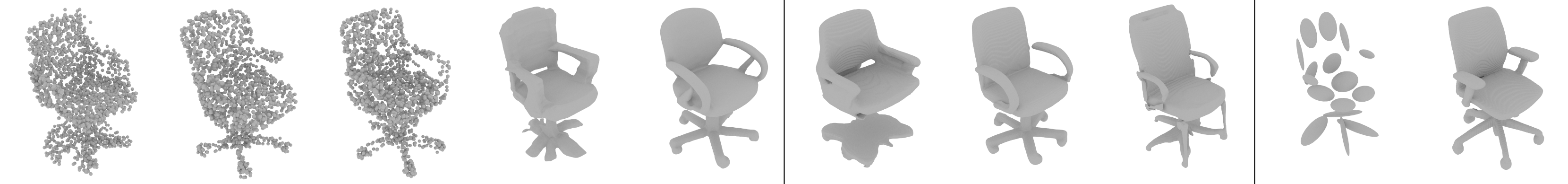}
\Image{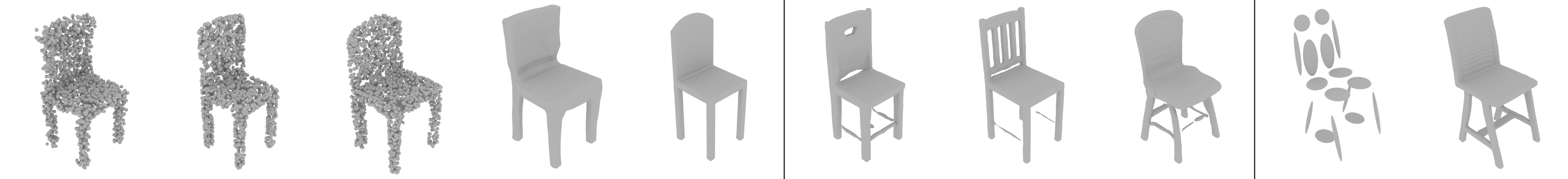}
\Image{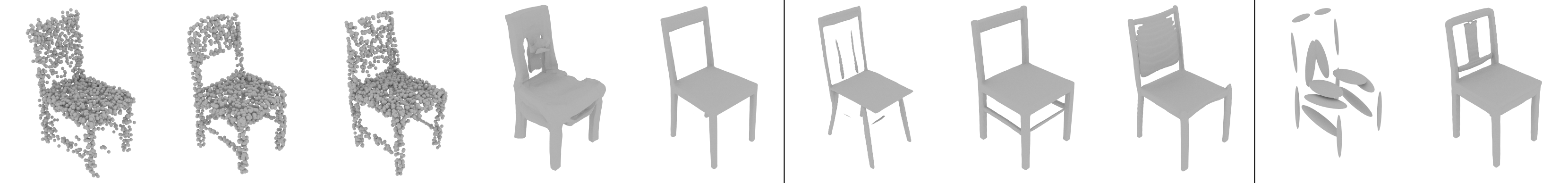}
\Image{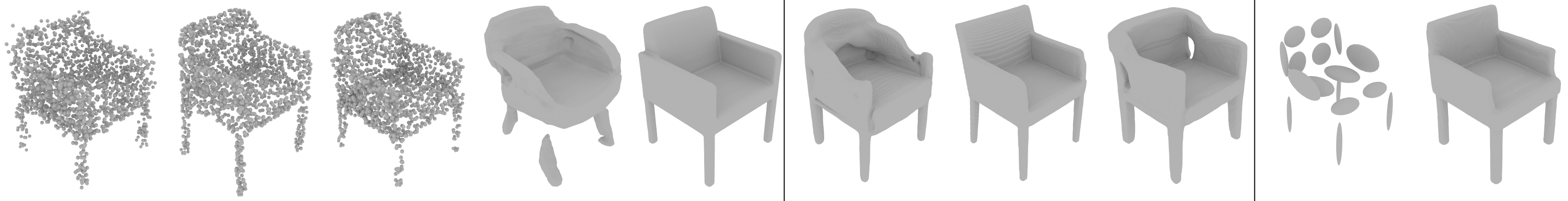}
\Image{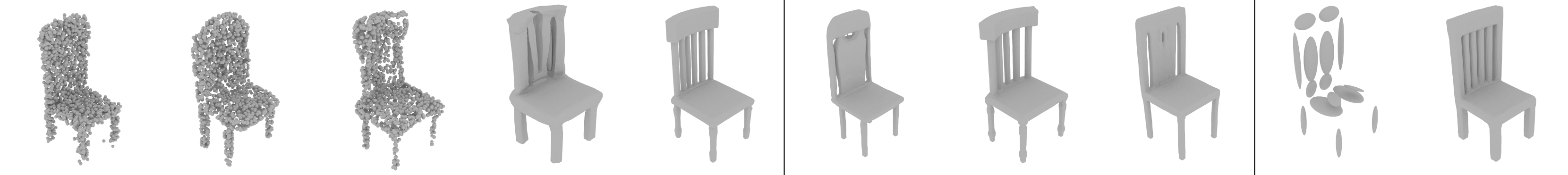}

\Image{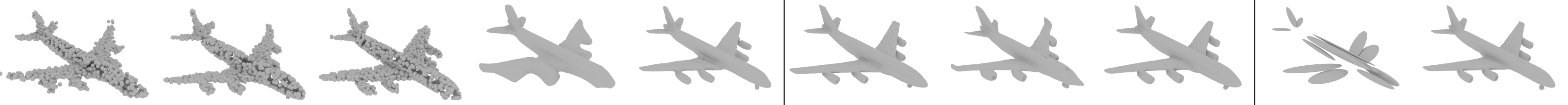}
\Image{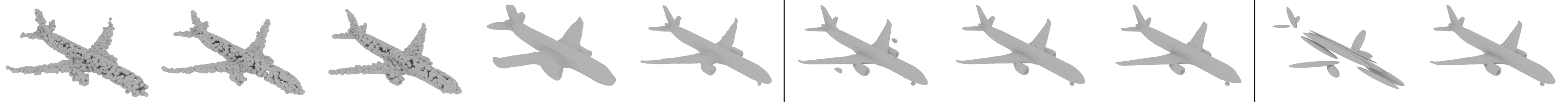}
\Image{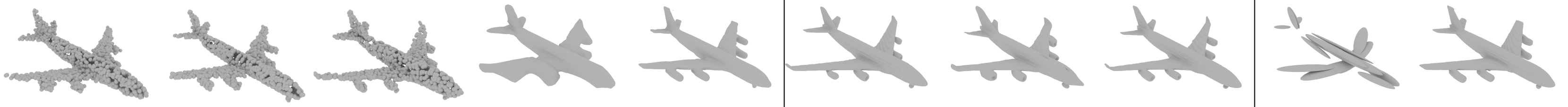}
\Image{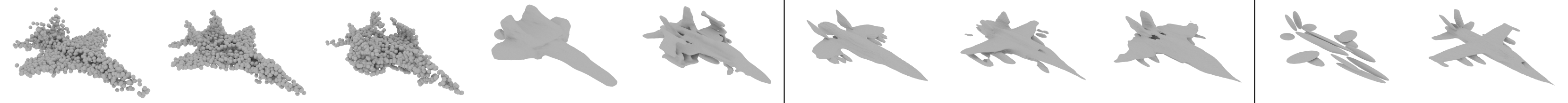}
\Image{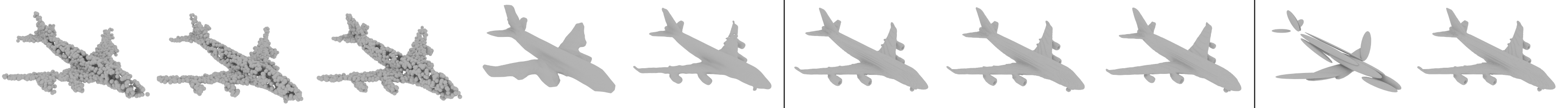}
\Image{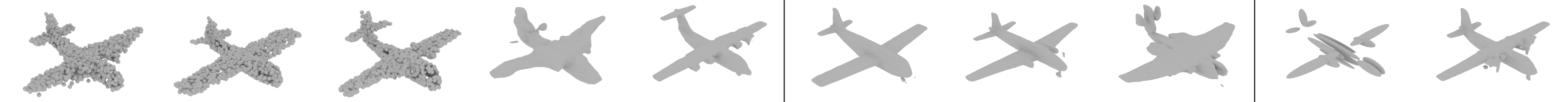}
\Image{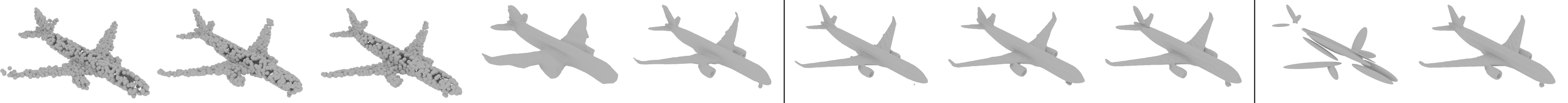}
\Image{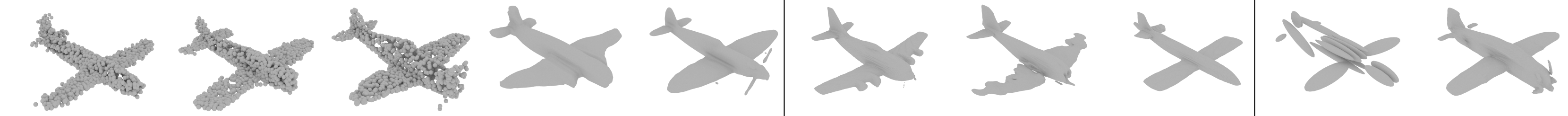}
\Image{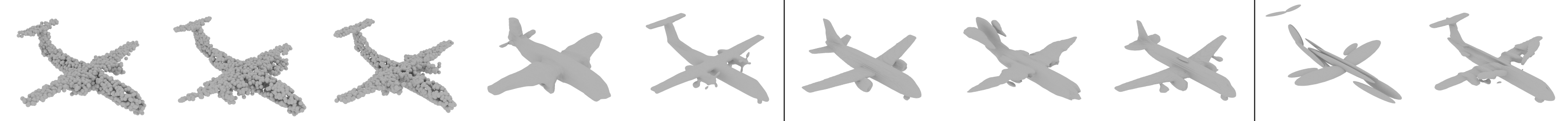}
\Image{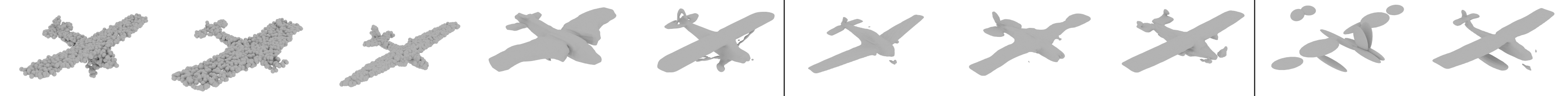}
\Image{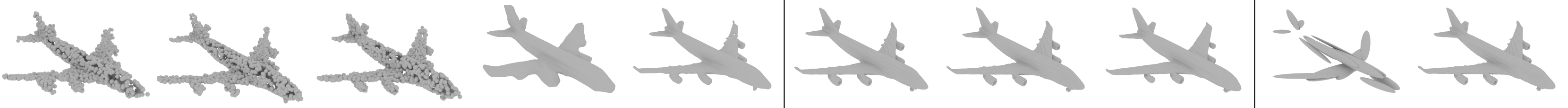}
\Image{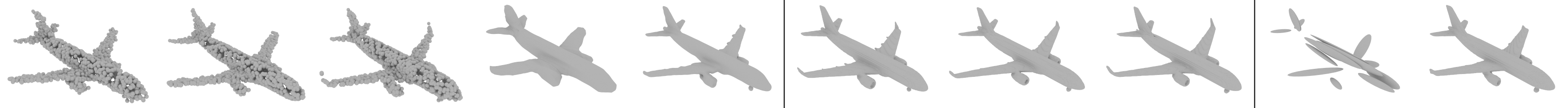}
\Image{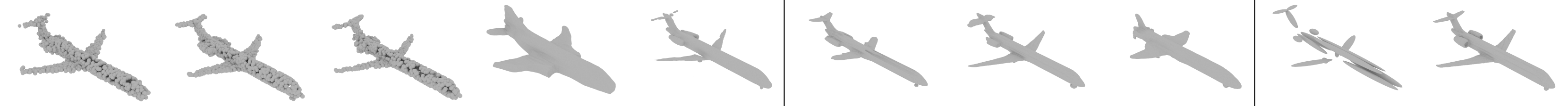}
\Image{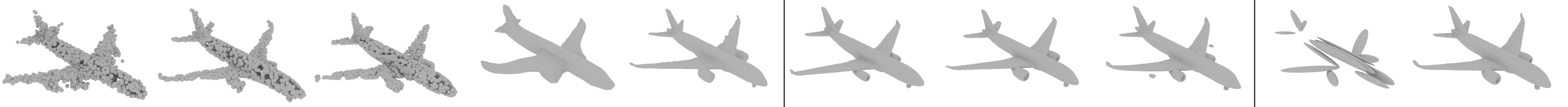}
\Image{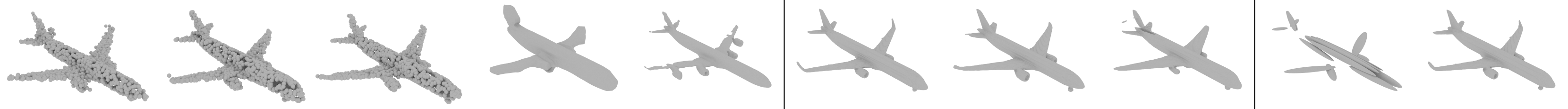}
\Image{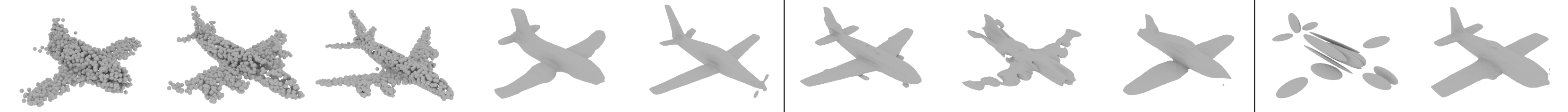}}

\makeatletter
\def\Image#1{%
  \multicolumn{\LT@cols}{l}{\includegraphics[width=\textwidth]{#1}}\\
}
\makeatother

\setlength{\tabcolsep}{0em}
\def\arraystretch{0.0}
\renewcommand\tabularxcolumn[1]{m{#1}}
\newcolumntype{Z}{>{\centering\arraybackslash}m{0.1\textwidth}}
{\scriptsize
\begin{longtable}{ZZZZZ|ZZZ|ZZ}
DPM~\cite{Luo:2021DPM} & PVD~\cite{Zhou:2021PVD} & LION~\cite{Zeng:2022LION} & Voxel-GAN~\cite{Kleineberg:2020VoxelGAN} & \makecell{Neural \\Wavelet\cite{Hui:2022NeuralWavelet}} & \makecell{SPAGHETTI\\\cite{Hertz:2022Spaghetti}} & \makecell{Diff. of\\$\B{z}$} & \makecell{Diff. of\\$\{\B{p}_i\}_{i=1}^N$} & Gaussians & \makecell{\salad{}\\(Ours)} \\
  \midrule
  \endhead

  \bottomrule
  \endfoot

  \Image{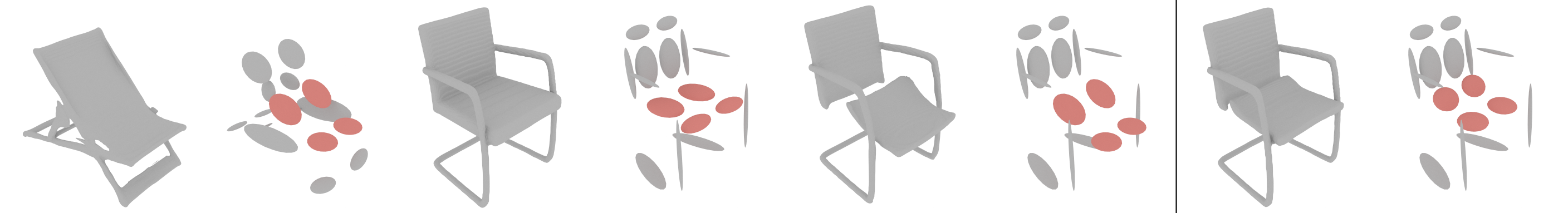}
\Image{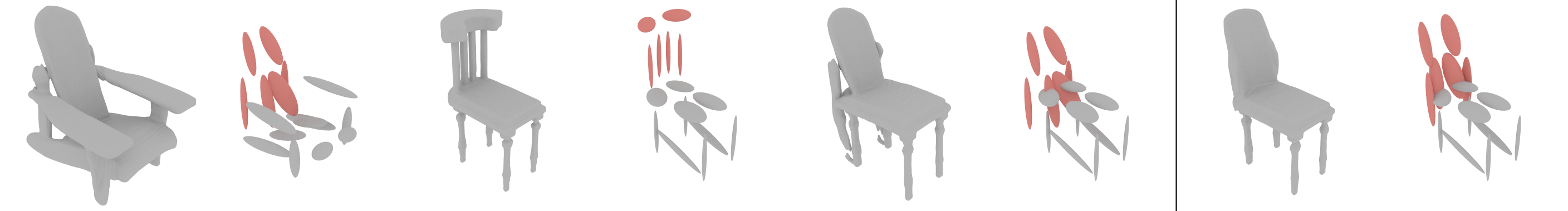}
\Image{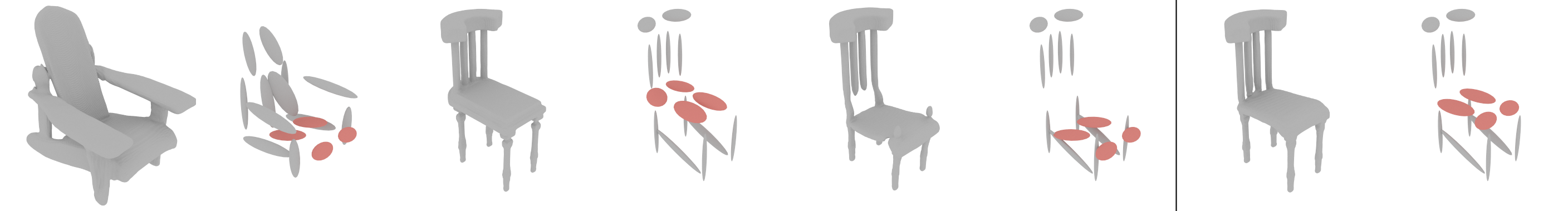}
\Image{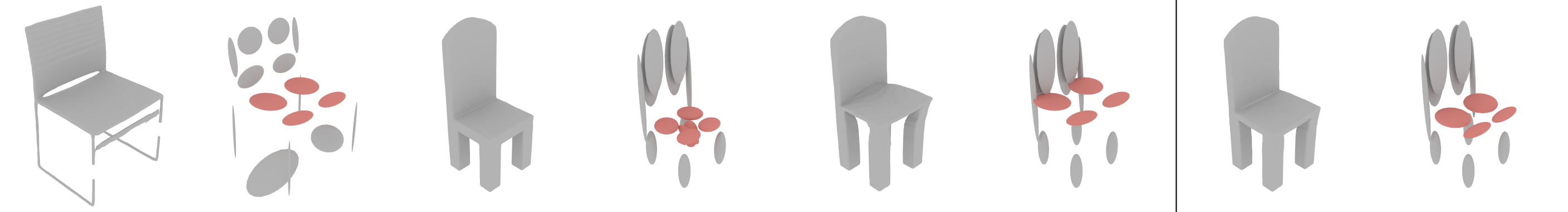}
\Image{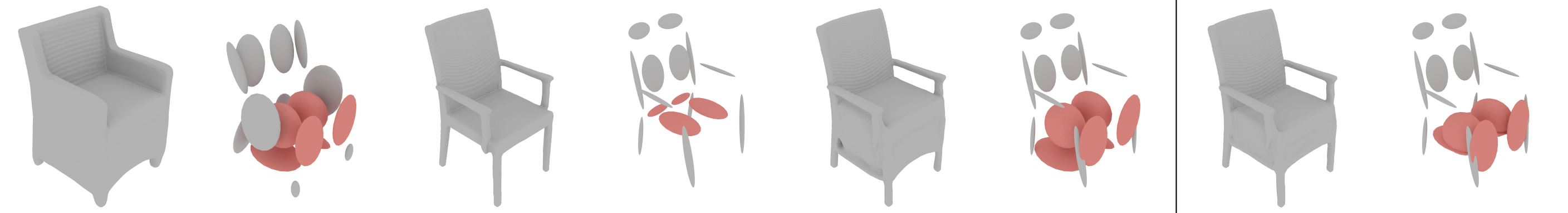}
\Image{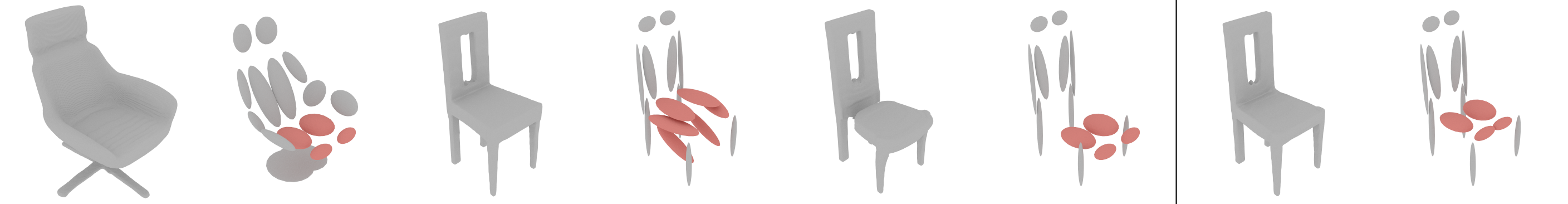}
\Image{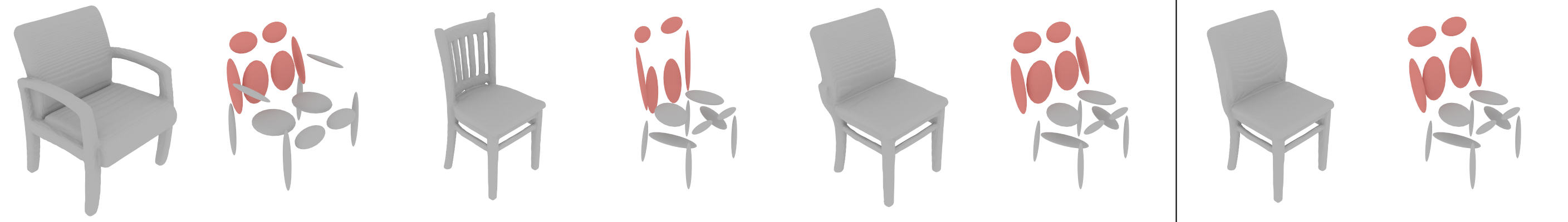}
\Image{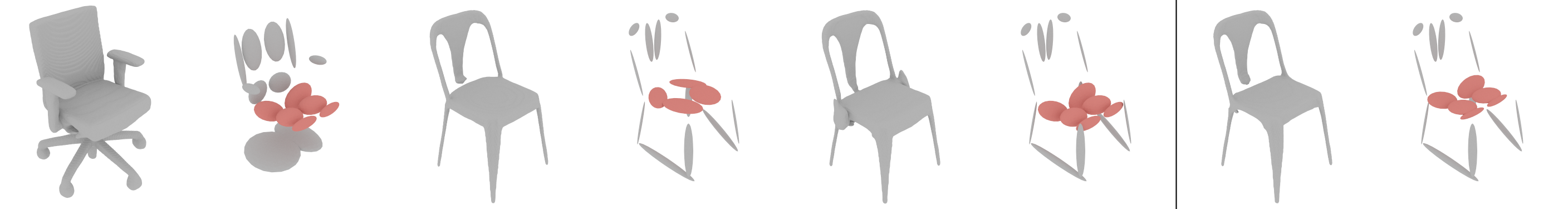}
\Image{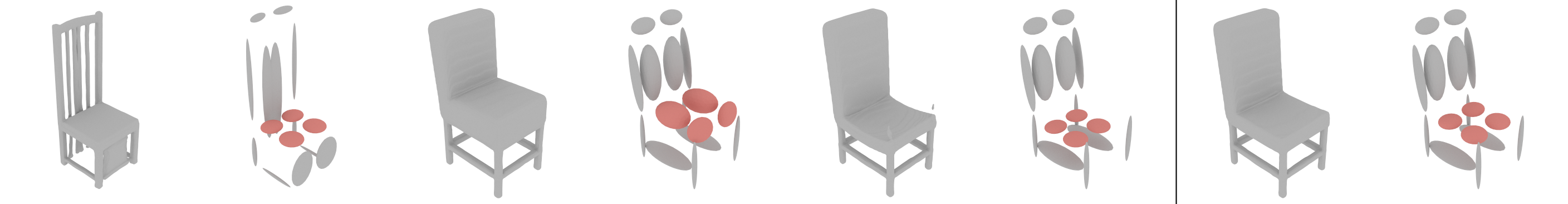}
\Image{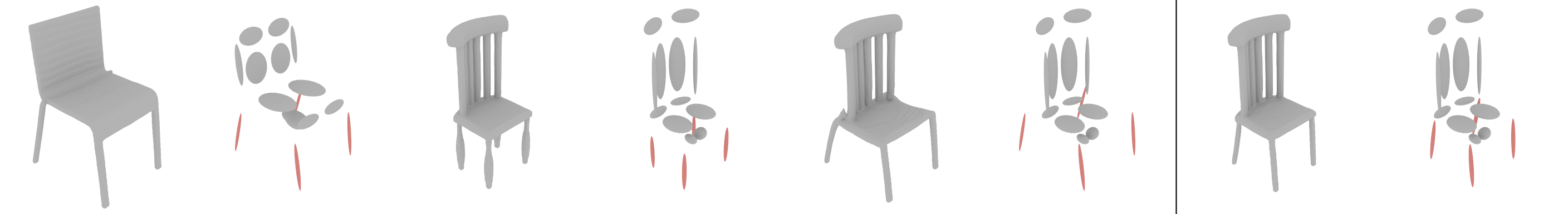}
\Image{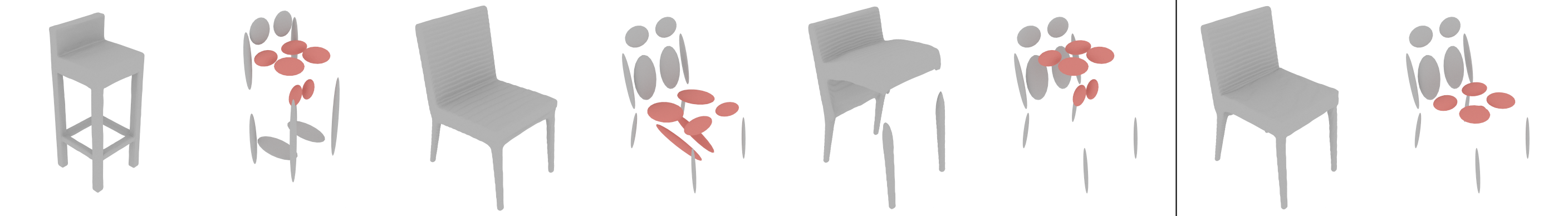}

\Image{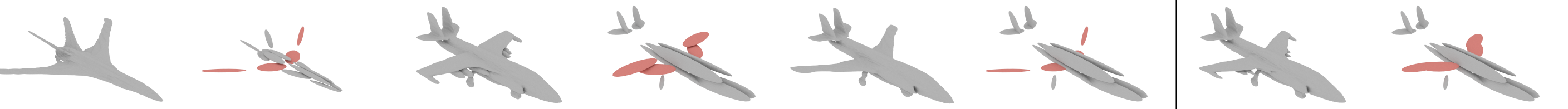}
\Image{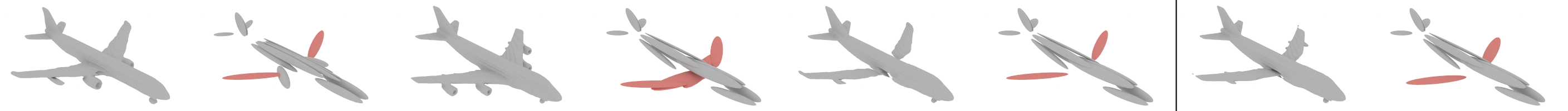}
\Image{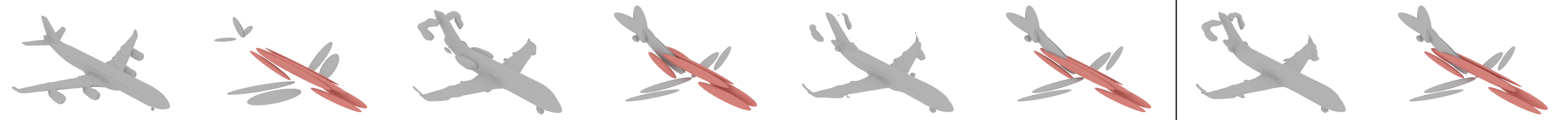}
\Image{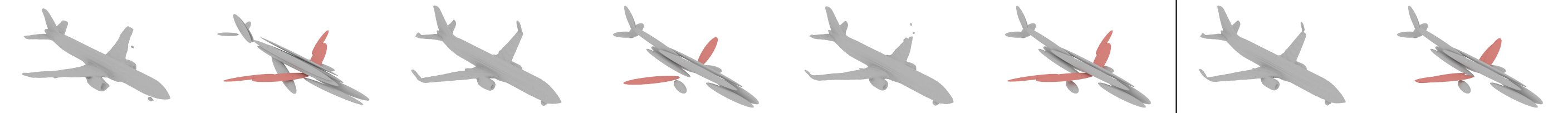}
\Image{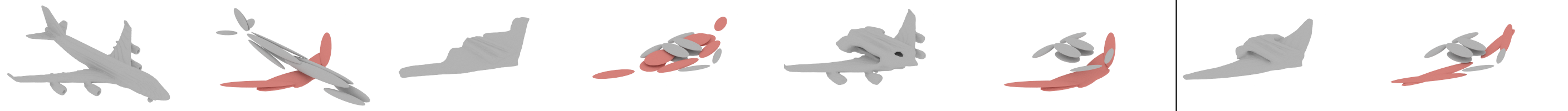}
\Image{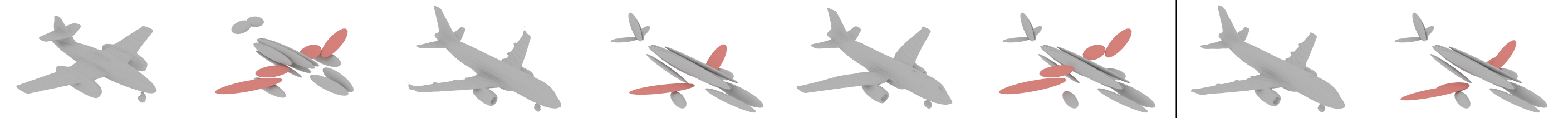}
\Image{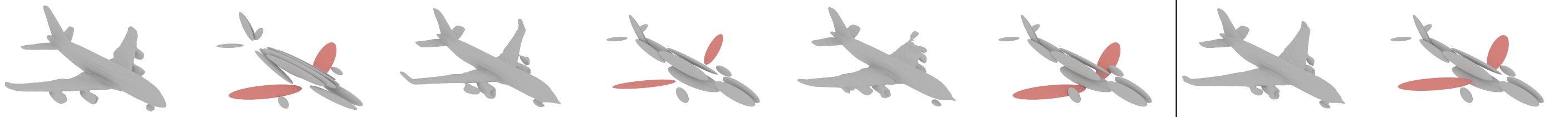}
\Image{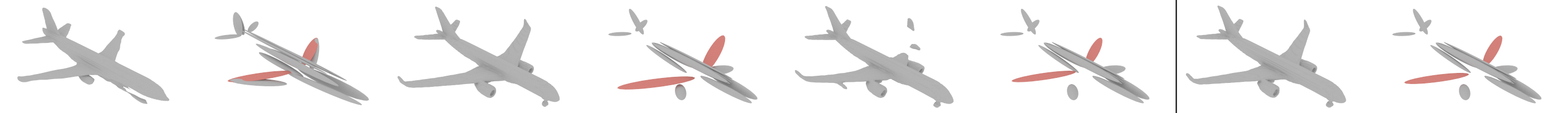}
\Image{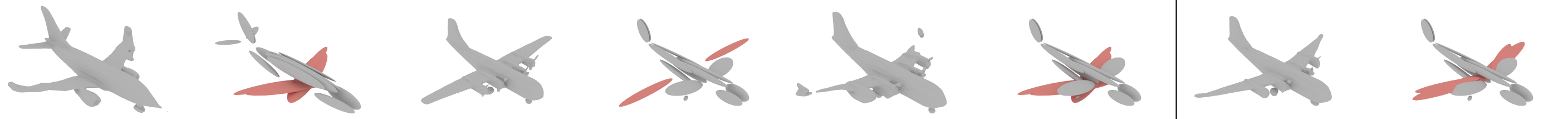}

\Image{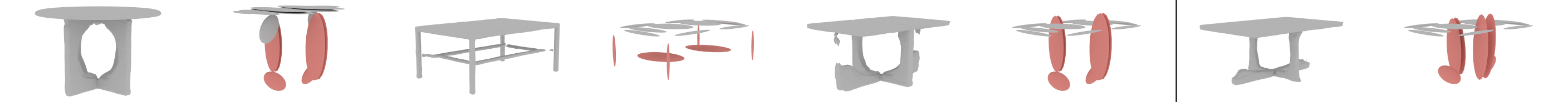}
\Image{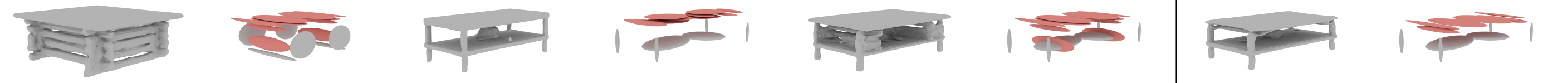}
\Image{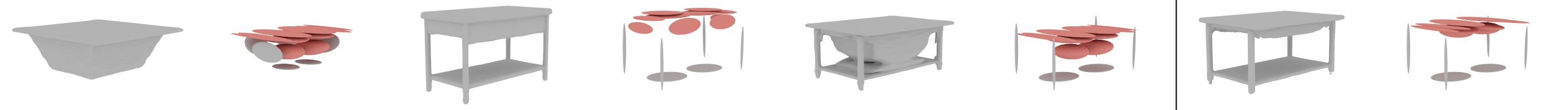}
\Image{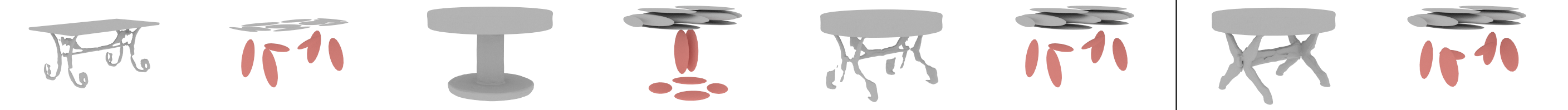}
\Image{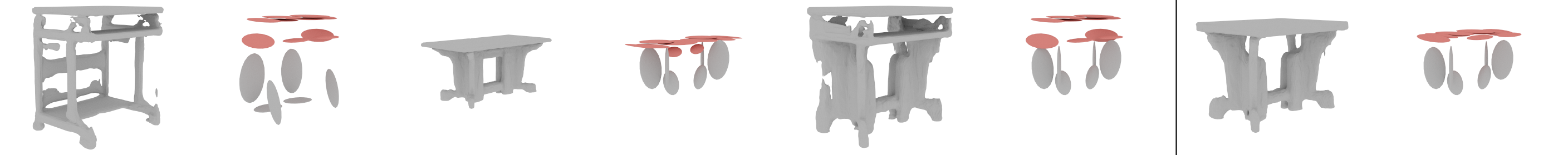}

\end{longtable}
}


\clearpage
\newpage

\subsection{More Qualitative Comparisons on Part Completion}
\label{sec:more_part_completion_comparison}
We report more qualitative comparisons on part completion with \emph{chair} and \emph{airplane} classes, as shown in Figure~\ref{fig:shape_completion}~\refinpaper{}. 

\input{figures/part_completion/supp_part_completion}


\clearpage
\newpage
\subsection{More Qualitative Results on Part Mixing and Refinement}
\label{sec:more_part_mixing}
We report more qualitative results on part mixing and refinement with \emph{chair}, \emph{airplane} and \emph{table} classes, as shown in Figure~\ref{fig:part_mixing}~\refinpaper{}.

\newcommand{\AllComparisonImages}{\Image{figures/part_mixing/chairs/a2b_0-1.png}
\Image{figures/part_mixing/chairs/a2b_24-0.png}
\Image{figures/part_mixing/chairs/a2b_24-1.png}
\Image{figures/part_mixing/chairs/a2b_3-1.png}
\Image{figures/part_mixing/chairs/a2b_6-1.png}
\Image{figures/part_mixing/chairs/a2b_9-1.png}
\Image{figures/part_mixing/chairs/a2b_8-0.png}
\Image{figures/part_mixing/chairs/a2b_21-1.png}
\Image{figures/part_mixing/chairs/a2b_63-1.png}
\Image{figures/part_mixing/chairs/a2b_69-2.png}
\Image{figures/part_mixing/chairs/a2b_79-1.png}

\Image{figures/part_mixing/airplanes/a2b_7-1.png}
\Image{figures/part_mixing/airplanes/a2b_10-1.png}
\Image{figures/part_mixing/airplanes/a2b_30-0.png}
\Image{figures/part_mixing/airplanes/a2b_35-1.png}
\Image{figures/part_mixing/airplanes/a2b_36-1.png}
\Image{figures/part_mixing/airplanes/a2b_41-1.png}
\Image{figures/part_mixing/airplanes/a2b_54-1.png}
\Image{figures/part_mixing/airplanes/b2a_8-1.png}
\Image{figures/part_mixing/airplanes/b2a_51-1.png}

\Image{figures/part_mixing/tables/a2b_3-1.png}
\Image{figures/part_mixing/tables/a2b_22-0.png}
\Image{figures/part_mixing/tables/a2b_38-0.png}
\Image{figures/part_mixing/tables/b2a_10-1.png}
\Image{figures/part_mixing/tables/b2a_14-0.png}
}

\makeatletter
\def\Image#1{%
  \multicolumn{\LT@cols}{l}{\includegraphics[width=0.8\textwidth]{#1}}\\
  \midrule 
}
\makeatother

\setlength{\tabcolsep}{0em}
\def\arraystretch{0.0}
\renewcommand\tabularxcolumn[1]{m{#1}}
\newcolumntype{W}{>{\centering\arraybackslash}m{0.2\textwidth}}

{\scriptsize
\begin{longtable}{WWW|W}
Shape A & Shape B & A$\rightarrow$B & \makecell{A$\rightarrow$B\\Refined} \\ 
  \midrule
  \endhead

  \endfoot

  \Image{figures/part_mixing/chairs/a2b_0-1.png}
\Image{figures/part_mixing/chairs/a2b_24-0.png}
\Image{figures/part_mixing/chairs/a2b_24-1.png}
\Image{figures/part_mixing/chairs/a2b_3-1.png}
\Image{figures/part_mixing/chairs/a2b_6-1.png}
\Image{figures/part_mixing/chairs/a2b_9-1.png}
\Image{figures/part_mixing/chairs/a2b_8-0.png}
\Image{figures/part_mixing/chairs/a2b_21-1.png}
\Image{figures/part_mixing/chairs/a2b_63-1.png}
\Image{figures/part_mixing/chairs/a2b_69-2.png}
\Image{figures/part_mixing/chairs/a2b_79-1.png}

\Image{figures/part_mixing/airplanes/a2b_7-1.png}
\Image{figures/part_mixing/airplanes/a2b_10-1.png}
\Image{figures/part_mixing/airplanes/a2b_30-0.png}
\Image{figures/part_mixing/airplanes/a2b_35-1.png}
\Image{figures/part_mixing/airplanes/a2b_36-1.png}
\Image{figures/part_mixing/airplanes/a2b_41-1.png}
\Image{figures/part_mixing/airplanes/a2b_54-1.png}
\Image{figures/part_mixing/airplanes/b2a_8-1.png}
\Image{figures/part_mixing/airplanes/b2a_51-1.png}

\Image{figures/part_mixing/tables/a2b_3-1.png}
\Image{figures/part_mixing/tables/a2b_22-0.png}
\Image{figures/part_mixing/tables/a2b_38-0.png}
\Image{figures/part_mixing/tables/b2a_10-1.png}
\Image{figures/part_mixing/tables/b2a_14-0.png}

\end{longtable}
}


\clearpage
\newpage

\subsection{More Qualitative Comparisons on Text-Guided Shape Generation}
\label{sec:more_text_generation}
We report more qualitative comparisons on text-guided shape generation between AutoSDF~\cite{Mittal:2022Autosdf} and \salad{}.

\begin{multicols}{2}
{
\footnotesize
\centering
\renewcommand\tabularxcolumn[1]{m{#1}}
\newcolumntype{Y}{>{\centering\arraybackslash}X}
\newcolumntype{Z}{>{\centering\arraybackslash}m{0.2\textwidth}}
\begin{tabularx}{0.35\textwidth}{Y | Y}
    \centering
    \scriptsize{AutoSDF~\cite{Mittal:2022Autosdf}} & \scriptsize{\salad{} (Ours)} \\
     \midrule
    \multicolumn{2}{c}{\includegraphics[width=0.2\textwidth]{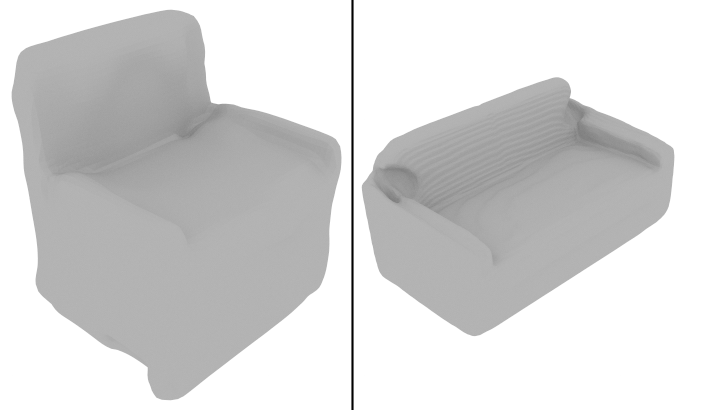}} \\ 
    \multicolumn{2}{c}{``\texttt{fat no legs.}''} \\
    \midrule

    \multicolumn{2}{c}{\includegraphics[width=0.2\textwidth]{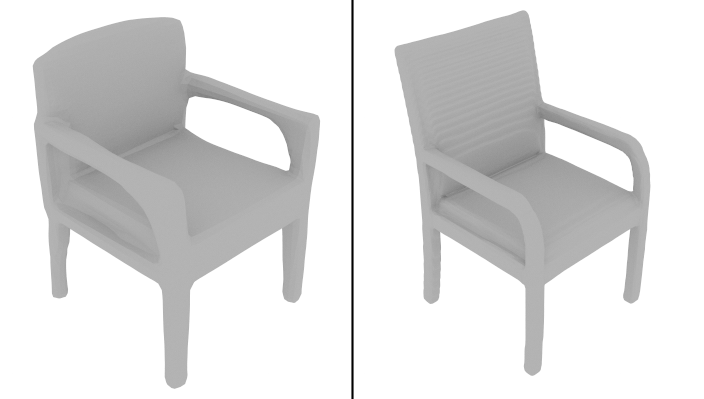}} \\ 
    \multicolumn{2}{c}{``\texttt{thin/skinny legs with chair arms.}''} \\
    \midrule

    \multicolumn{2}{c}{\includegraphics[width=0.2\textwidth]{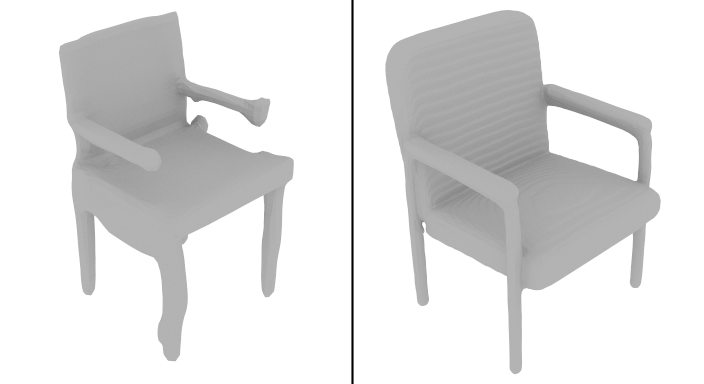}} \\ 
    \multicolumn{2}{c}{``\texttt{the target has very tiny arms.}''} \\
    \midrule

    \multicolumn{2}{c}{\includegraphics[width=0.2\textwidth]{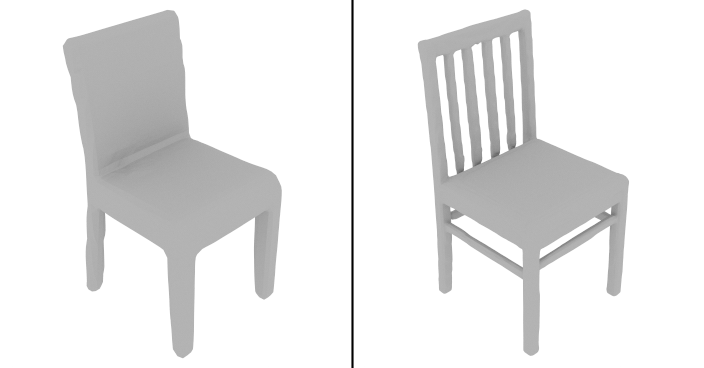}} \\ 
    \multicolumn{2}{c}{``\texttt{with a narrow slat across my back.}''} \\
    \midrule

    \multicolumn{2}{c}{\includegraphics[width=0.2\textwidth]{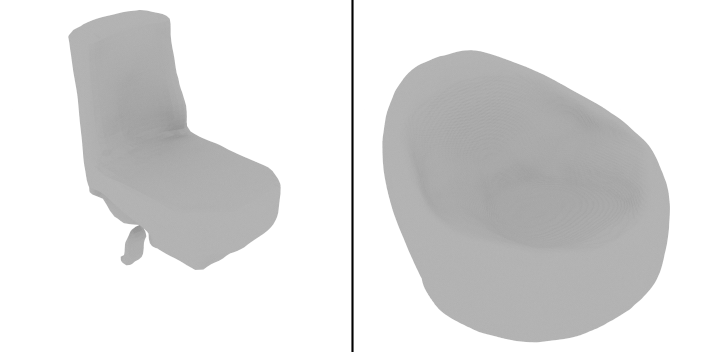}} \\ 
    \multicolumn{2}{c}{``\texttt{round chair with round back.}''} \\
    \midrule

    \multicolumn{2}{c}{\includegraphics[width=0.2\textwidth]{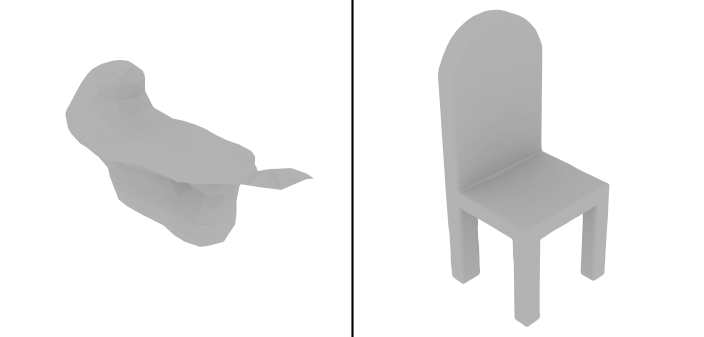}} \\ 
    \multicolumn{2}{c}{``\texttt{curved top.}''} \\
    \midrule

    \multicolumn{2}{c}{\includegraphics[width=0.2\textwidth]{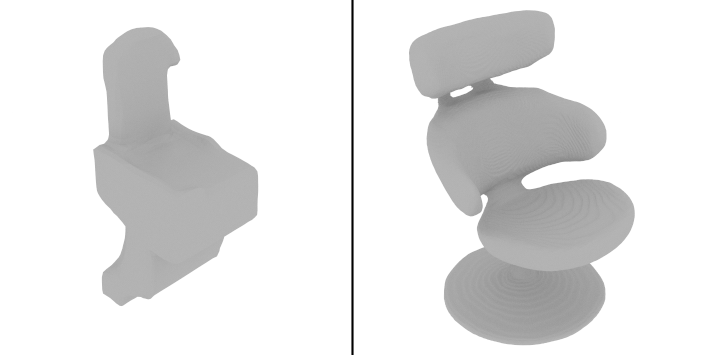}} \\ 
    \multicolumn{2}{c}{``\texttt{oval footrest.}''} \\
    \midrule

    \multicolumn{2}{c}{\includegraphics[width=0.2\textwidth]{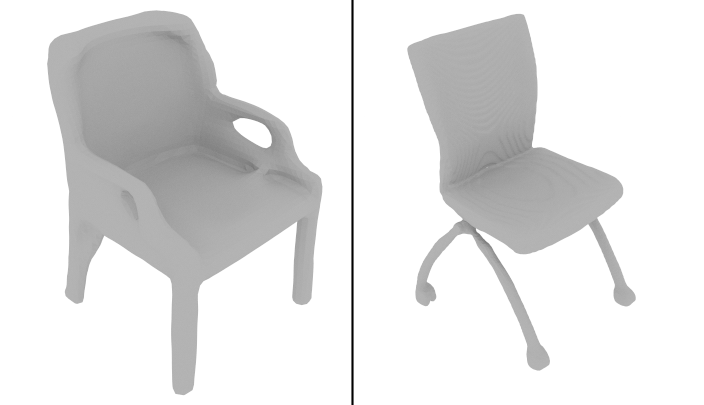}} \\ 
    \multicolumn{2}{c}{\makecell{``\texttt{wrap around curved back}\\\texttt{narrow legs.}''}} \\
    \midrule

    \multicolumn{2}{c}{\includegraphics[width=0.2\textwidth]{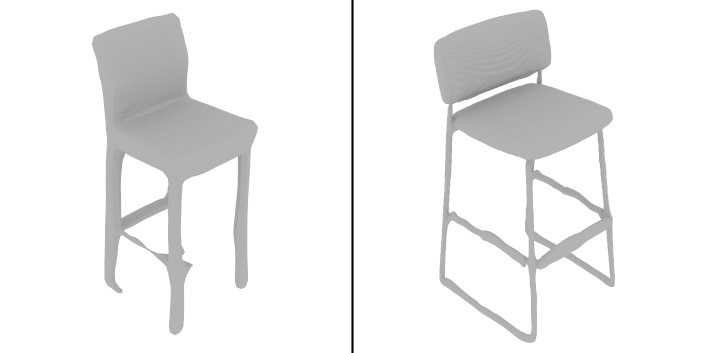}} \\ 
    \multicolumn{2}{c}{\makecell{``\texttt{this chair is very tall with}\\\texttt{skinny legs on it.}''}} \\
    \midrule

\label{fig:supp_text_generation}
\end{tabularx}
}
\columnbreak

{
\footnotesize
\centering
\renewcommand\tabularxcolumn[1]{m{#1}}
\newcolumntype{Y}{>{\centering\arraybackslash}X}
\newcolumntype{Z}{>{\centering\arraybackslash}m{0.2\textwidth}}
\begin{tabularx}{0.35\textwidth}{Y | Y}
    \scriptsize{
     AutoSDF~\cite{Mittal:2022Autosdf}} & \scriptsize{\salad{} (Ours)} \\
     \midrule
    \multicolumn{2}{c}{\includegraphics[width=0.2\textwidth]{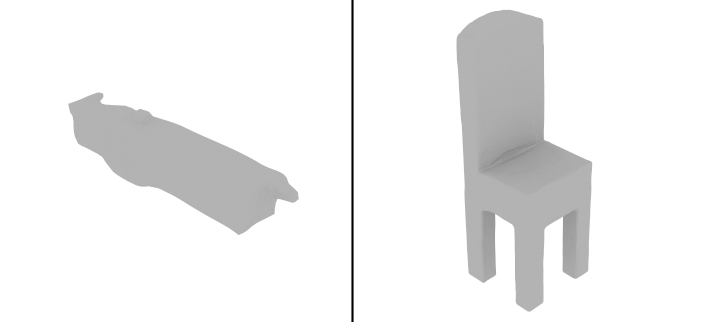}} \\ 
    \multicolumn{2}{c}{``\texttt{curved solid back.}''} \\
    \midrule
    
    \multicolumn{2}{c}{\includegraphics[width=0.2\textwidth]{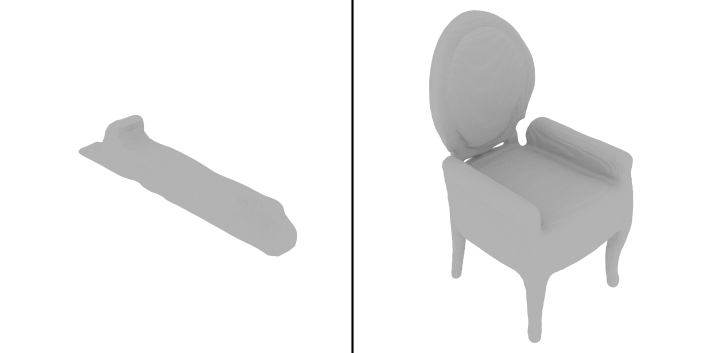}} \\ 
    \multicolumn{2}{c}{\makecell{``\texttt{rounded back.}''}} \\
    \midrule
    
    \multicolumn{2}{c}{\includegraphics[width=0.2\textwidth]{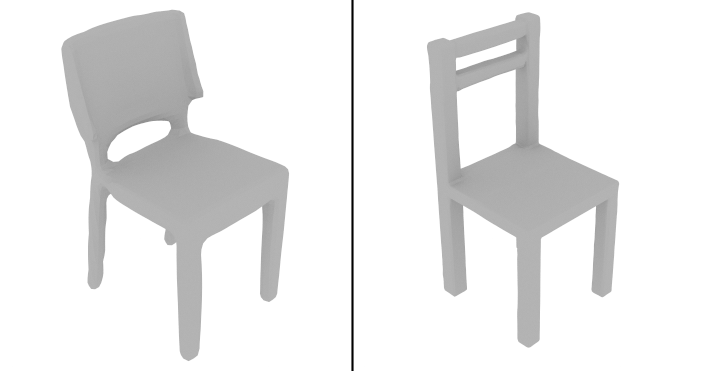}} \\ 
    \multicolumn{2}{c}{\makecell{``\texttt{has an opening in the back}\\\texttt{of the chair.}''}} \\
    \midrule
    
    \multicolumn{2}{c}{\includegraphics[width=0.2\textwidth]{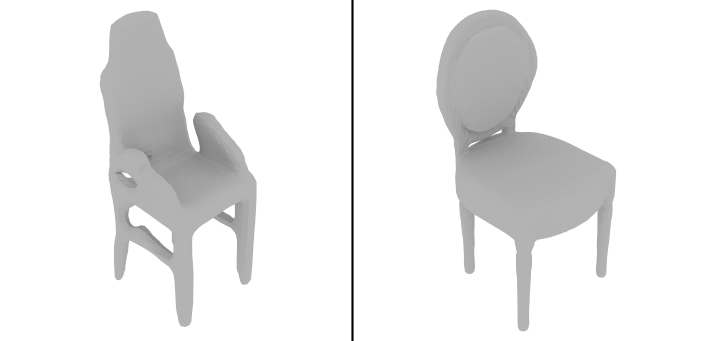}} \\ 
    \multicolumn{2}{c}{\makecell{``\texttt{the one with the oval}\\\texttt{shaped back.}''}} \\
    \midrule

    \multicolumn{2}{c}{\includegraphics[width=0.2\textwidth]{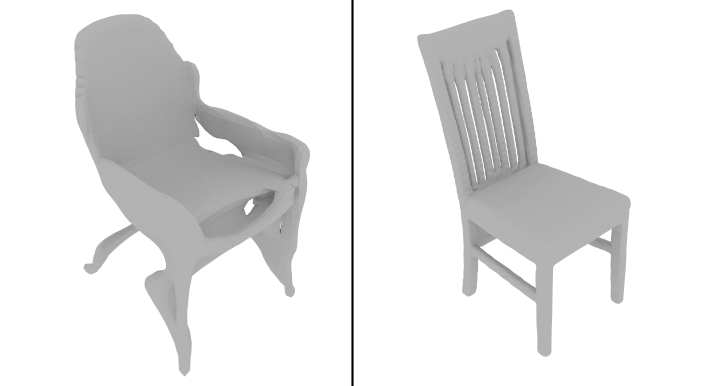}} \\ 
    \multicolumn{2}{c}{\makecell{``\texttt{the one that look most like}\\\texttt{a lawn chair. net-like back.}''}} \\
    \midrule

    \multicolumn{2}{c}{\includegraphics[width=0.2\textwidth]{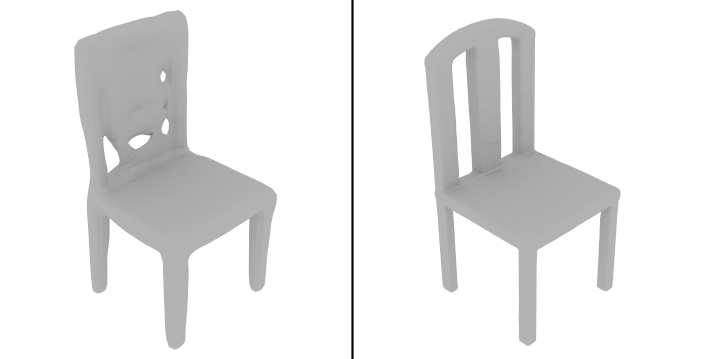}} \\ 
    \multicolumn{2}{c}{\makecell{``\texttt{dining room chair with}\\\texttt{fancy holes in back.}''}} \\
    \midrule


    \multicolumn{2}{c}{\includegraphics[width=0.2\textwidth]{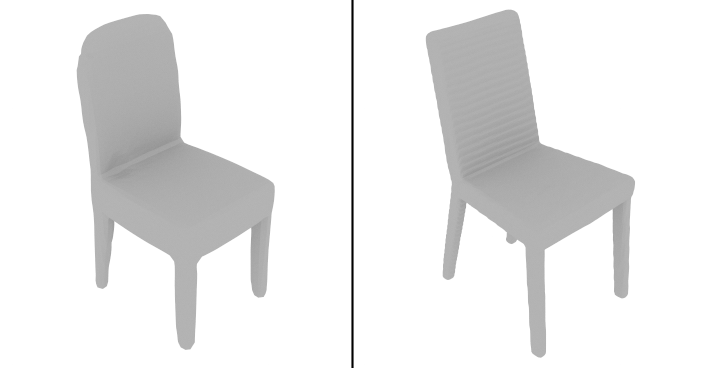}} \\ 
    \multicolumn{2}{c}{\makecell{``\texttt{regular looking back, no arms.}''}} \\
    \midrule

    \multicolumn{2}{c}{\includegraphics[width=0.2\textwidth]{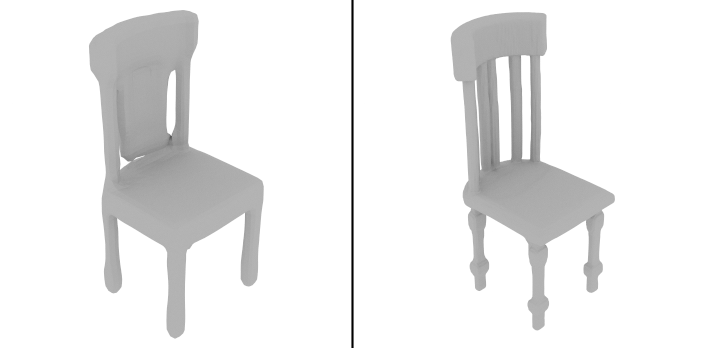}} \\ 
    \multicolumn{2}{c}{\makecell{``\texttt{5 lines, with curve.}''}} \\    
    \midrule

\label{fig:supp_text_generation}
\end{tabularx}
}
\end{multicols}

\clearpage
\newpage
\onecolumn
\subsection{More Qualitative Results on Text-Guided Part Completion}
\label{sec:more_text_completion}
We report more qualitative results on text-guided part completion leveraging \salad{} and \gaussglot{}. In the figure below, the parts selected by \gaussglot{} from the text are highlighted by red. Text-conditioned \salad{} completes the selected parts to match the text via the guided reverse process.

{
\footnotesize
\centering
\renewcommand\tabularxcolumn[1]{m{#1}}
\newcolumntype{Y}{>{\centering\arraybackslash}X}
\newcolumntype{Z}{>{\centering\arraybackslash}m{0.125\textwidth}}
\begin{longtable}{ZZZZ|ZZZZ}
     \scriptsize{Input Mesh} & \scriptsize{\makecell{Input\\Gaussians}} & \scriptsize{Output Mesh} & \scriptsize{\makecell{Output\\Gaussians}} & \scriptsize{Input Mesh} & \scriptsize{\makecell{Input\\Gaussians}} & \scriptsize{Output Mesh} & \scriptsize{\makecell{Output\\Gaussians}} \\
     \midrule
    \endhead
  \endfoot
    \multicolumn{8}{c}{\includegraphics[width=\textwidth]{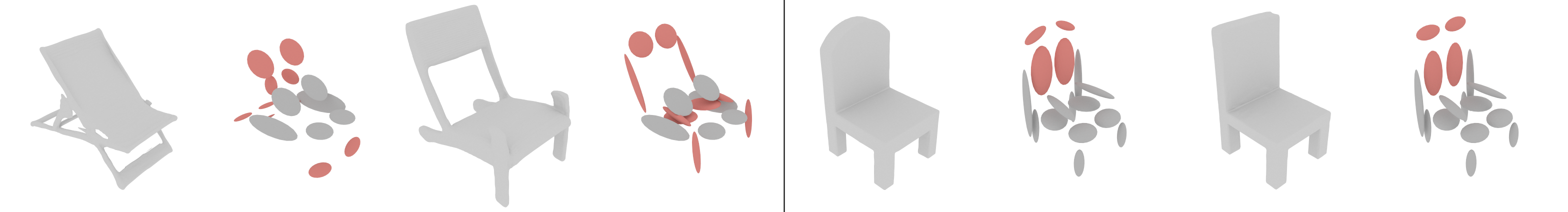}} \\
\multicolumn{4}{c}{``\texttt{four legs and a straight back}''} & \multicolumn{4}{c}{``\texttt{straight rectangular back}''} \\
\midrule

\multicolumn{8}{c}{\includegraphics[width=\textwidth]{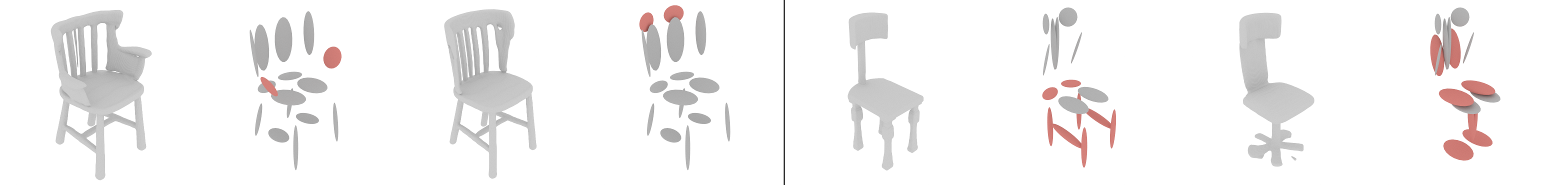}} \\
\multicolumn{4}{c}{``\texttt{chair with no arms}''} & \multicolumn{4}{c}{``\texttt{swivel legs}''} \\
\midrule

\multicolumn{8}{c}{\includegraphics[width=\textwidth]{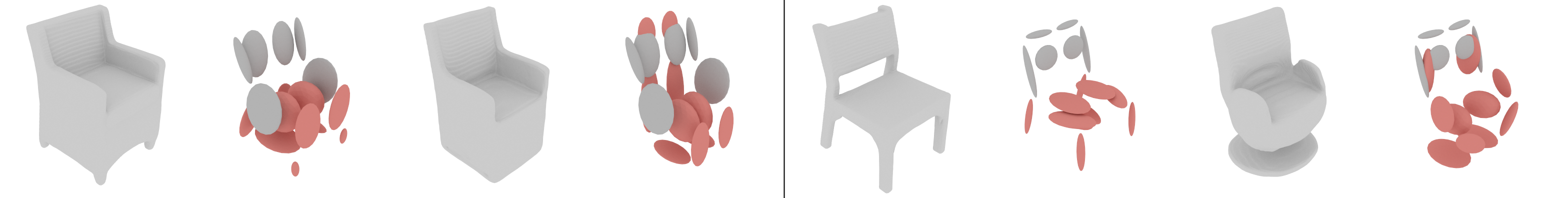}} \\
\multicolumn{4}{c}{``\texttt{solid base and no leg}''} & \multicolumn{4}{c}{``\texttt{round seat has arms and a circle base}''} \\
\midrule

\multicolumn{8}{c}{\includegraphics[width=\textwidth]{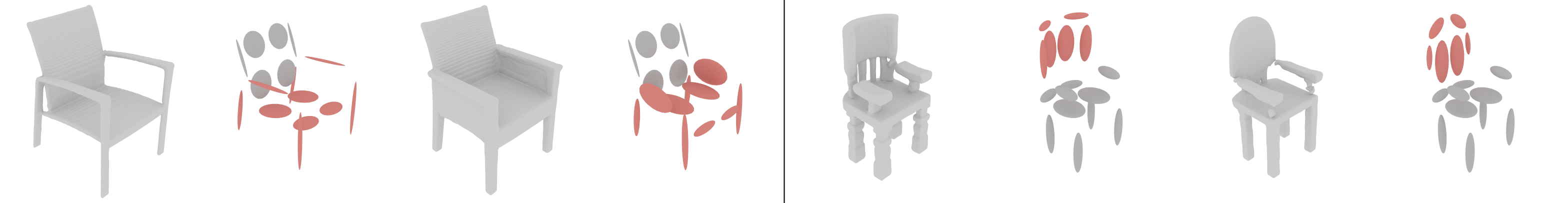}} \\
\multicolumn{4}{c}{``\texttt{thick legs and arms}''} & \multicolumn{4}{c}{``\texttt{circular back}''} \\
\midrule

\multicolumn{8}{c}{\includegraphics[width=\textwidth]{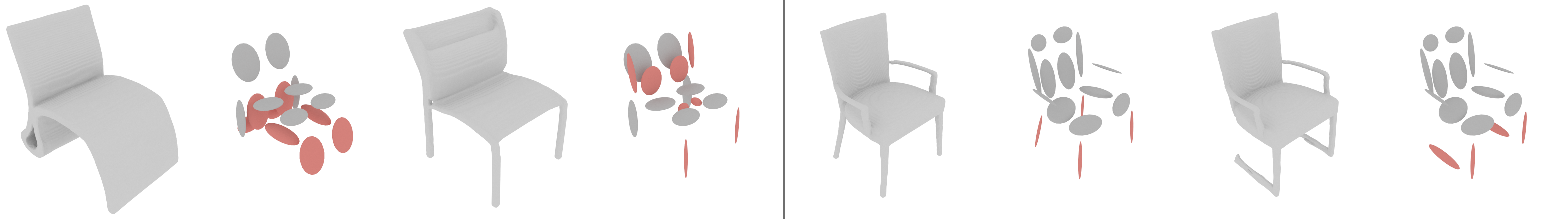}} \\
\multicolumn{4}{c}{``\texttt{four thin legs}''} & \multicolumn{4}{c}{``\texttt{it only has two legs}''} \\
\midrule

\label{fig:more_text_completion}

\end{longtable}
}

\clearpage
\newpage
\twocolumn

\else
    
\fi

\end{document}